\documentclass[preprint,authoryear,12pt]{elsarticle}

\usepackage{amssymb}
\usepackage{fullpage}
\usepackage{graphicx}
\usepackage{amsmath}
\usepackage{amsthm}
\usepackage{eucal}
\usepackage{amssymb}
\usepackage{mathrsfs}
\usepackage{lscape}
\usepackage{multirow}
\usepackage{array}
\usepackage[table]{xcolor}
\usepackage{pdflscape}
\usepackage{lineno}
\usepackage{booktabs}
\usepackage[para,online,flushleft]{threeparttable}
\usepackage{caption} 
\usepackage{algorithm}
\usepackage{algpseudocode}

\newcommand{\argmin}{\operatornamewithlimits{argmin}}
\newcommand{\REV}[1]{{#1}}

\journal{Medical Image Analysis}

\begin{document}

\begin{frontmatter}



\title{Slice-to-volume medical image registration: a survey}


\author[1,2]{Enzo Ferrante}
\ead{e.ferrante@imperial.ac.uk}
\author[1,3]{Nikos Paragios}
\ead{nikos.paragios@ecp.fr}
\address[1]{Center for Visual Computing, CentraleSupelec, INRIA, Universite Paris-Saclay. \\Grande Voie des Vignes, Chatenay-Malabry, 92295, France}
\address[2]{Biomedical Image Analysis (BioMedIA) Group, Department of Computing, Imperial College London. \\South Kensington Campus, 180 Queen's Gate, London, SW7 2AZ, UK}
\address[3]{TheraPanacea, 24 Rue du Faubourg Saint-Jacques, 75014 Paris}

\begin{abstract}
During the last decades, the research community of medical imaging has witnessed continuous advances in image registration methods, which pushed the limits of the state-of-the-art and enabled the development of novel medical procedures. A particular type of image registration problem, known as slice-to-volume registration, played a fundamental role in areas like image guided surgeries and volumetric image reconstruction. However, to date, and despite the extensive literature available on this topic, no survey has been written to discuss this challenging problem. This paper introduces the first comprehensive survey of the literature about slice-to-volume registration, presenting a categorical study of the algorithms according to an ad-hoc taxonomy and analyzing advantages and disadvantages of every category. We draw some general conclusions from this analysis and present our perspectives on the future of the field.
\end{abstract}

\end{frontmatter}


%

\section{Introduction}
Image registration is the process of aligning and combining data coming from more than one image source into a unique coordinate system. This problem has become one of the pillars of computer vision and medical imaging. Slice-to-volume registration, a particular case of image registration problem, has received further attention in the medical imaging community during the last decade. In this case, instead of registering images with same dimension, we seek to determine the slice (corresponding to an arbitrary plane) from a given 3D volume that corresponds to an input 2D image. 

Several medical applications requiring slice-to-volume mapping have emerged and pushed the community towards developing more accurate and efficient strategies. Such medical imaging tasks can be classified in two main categories: those related to image fusion for image guided interventions; and those related to motion correction and volume reconstruction. In the first category, pre-operative 3D images and intra-operative 2D images need to be fused to guide surgeons during medical interventions. Slice-to-volume registration plays a key role in this process, allowing the physicians to navigate 3D pre-operative high-resolution annotated data using low-resolution 2D images acquired in real-time during surgery. In the second category, the goal is to correct for misaligned slices when reconstructing a volume of a certain modality. A typical approach to solve this task consists in mapping individual slices onto a reference volume in order to correct for inter-slice misalignment. Again, the development of accurate slice-to-volume registration algorithms is crucial to successfully tackle this problem.

Slice-to-volume registration is also known as 2D/3D registration, due to the dimension of the images involved in the registration process. However, this term is ambiguous since it describes two different problems depending on the technology used to capture the 2D image: it may be a projective (e.g. x-ray) or sliced/tomographic (e.g. US) image. Even if both problems share similarities in terms of image dimensionality, every formulation requires a different strategy to estimate the solution. The lack of perspective and different image geometry \citep{Birkfellner2007} inherent to both modalities, make it necessary to come up with distinct strategies to solve these registration problems. Moreover, a pixel in any 2D projective image does not correspond to only one voxel from the volume (this is the case for slice-to-volume), but to a projection of a set of them in certain perspective. Therefore, the type of functions used to measure similarities between the images is necessarily different in every case. While most of the projective 2D/3D image registration methods require to bring the images into dimensional correspondence (by different strategies like projection, back-projection or reconstruction \citep{Markelj2010}), in case of slice-to-volume registration, pixels from the 2D image can be directly compared with voxels from the volume. In this review we focus on slice-to-volume registration, while a more comprehensive overview about projective 2D to 3D image registration is presented \REV{in the survey by} \citep{Markelj2010}.

Slice-to-volume registration could be considered as an extreme case of 3D-3D registration, where one of the 3D images contains only one slice. Even if theoretically true, 3D-3D registration methods can not be extrapolated in a straightforward way to the slice-to-volume scenario. This holds, particularly, for registration methods based on image information, since the descriptors used to quantify similarities between images, normally assume that the amount of information available from both images is balanced. The fact that a single slice (or even a few sparse slices) provides less information than an entire volume, should be explicitly considered in the problem formulation. Moreover, specific geometrical constraints like planarity satisfaction and in-plane deformation restrictions, arise in the case of slice-to-volume registration, which are not applicable in the setting of dimensional correspondence.

\begin{figure}[t!]
  \centering
     \includegraphics[scale=0.88]{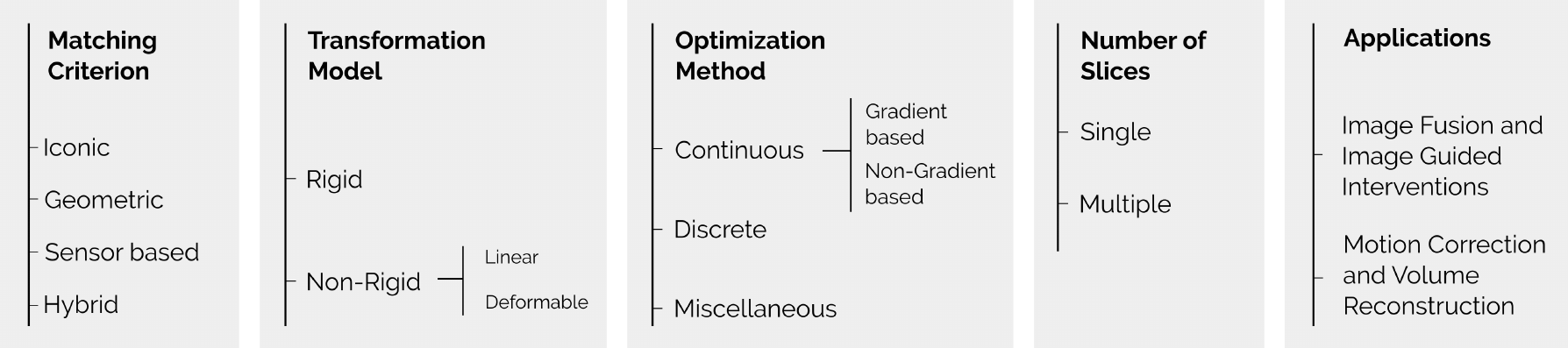}
     \caption{Taxonomy used to classify the slice-to-volume registration publications analyzed in this survey.}
  \label{fig:taxonomy}
\end{figure}

Despite the extensive literature available about slice-to-volume registration, to date, no survey has been written to discuss this emerging field. In this review, we discuss the literature about slice-to-volume registration, offering a comprehensive survey of the articles tackling this challenging problem, proposing a categorical study of the related algorithms according to a taxonomy and analyzing the advantages and disadvantages of each category. We propose a taxonomy based on previous surveys on image registration \citep{Markelj2010, Sotiras2013} and adapted to the particular slice-to-volume case (see Figure~\ref{fig:taxonomy}). We classified the related work according to several principles: (i) matching criterion (section~\ref{sec:chap3:matchingCriterion}), which specifies the strategies to define similarities between the images; (ii) transformation model (section~\ref{sec:chap3:transformationModel}), indicating the nature of the models used to bring images into spatial correspondence; (iii) optimization (section~\ref{sec:chap3:optimizationMethod}), differentiating the approaches according to the strategy used to infer the optimal transformation model; (iv) number of slices (section~\ref{sec:chap3:numberOfSlices}), which splits the methods in two groups, according to whether they require a single or multiple source image slices; (v) applications (section~\ref{sec:chap3:application}), where we identify the main clinical scenarios that have motivated research in the field; \REV{and finally (vi) validation strategy (section~\ref{sec:validation}) where we summarize the alternative  approaches adopted to validate the proposed slice-to-volume registration methods.}

Papers included in this review were systematically selected, searching for papers containing the term 'slice to volume' and 'registration' (and their variations), and carefully choosing those that either present a methodological contribution or apply slice-to-volume registration to a particular problem introducing some degree of novelty. 

\subsection{Definition of Slice-to-Volume Registration}
\label{sec:chap3:formalDefinition}
Let us start by giving a formal definition of slice-to-volume registration. Given a 2D image $I$ and a 3D volume $J$, we seek a mapping function $\hat{\varTheta}$ which optimally aligns the tomographic slice $I$ with the volumetric image $J$, through the minimization of the following objective function:

\begin{equation}
	\hat{\varTheta} = \argmin_{\varTheta} \mathcal{M}(I, J; \varTheta) + \mathcal{R}(\varTheta),
	\label{eq:chap3:generalSliceToVolume}
\end{equation}	

\noindent where $\mathcal{M}$ represents the image similarity term (the \REV{so-called} matching criterion) and $\mathcal{R}$ the regularization term. Note that this mapping may be rigid or non-rigid, depending on whether we allow image $I$ (or its corresponding reformatted slice from $J$) to be deformed or not. If we estimate only a rigid mapping (i.e. we calculate a 6 degrees of freedom rigid transformation or even a more restrictive one), we name the problem rigid slice-to-volume registration. In case we also infer some sort of deformation model or we consider more expressive linear transformations (such as affine transformations), we call it non-rigid registration. We refer the reader to section \ref{sec:chap3:transformationModel} for more information about different transformation models.  

\REV{The} matching criterion $\mathcal{M}$ measures the similarity between the 2D image and its corresponding mapping (slice) to the 3D volume. Usually, it is defined using intensity information or salient structures from $I$ and $J$. A complete discussion about matching criteria in the context of slice-to-volume registration (where we also consider methods that do not use any image information but rely on external sensors) is presented in section \ref{sec:chap3:matchingCriterion}. 

The regularization term $\mathcal{R}$ imposes constraints on the solution that can be used to render the problem well posed. It also may encode geometric properties on the extended (plane selection and plane deformation in case of non-rigid registration) transformation model. The choice of regularizer depends on the transformation model. While simple models like rigid body transformations can be explicitly estimated even without regularizer, the term $\mathcal{R}$ becomes crucial in more complex non-rigid scenarios to ensure realistic results. \REV{In the context of slice-to-volume registration, the regularizer can be used to impose planarity constraints to the solution (when out-of-plane deformations are not allowed) or to limit the out-of-plane deformation magnitude guaranteeing realistic and plausible transformations. When available, prior knowledge about tissue elasticity can also be encoded through the regularizer. Moreover, as we will see in Section \ref{sec:chap3:numberOfSlices}, when dealing with multiple slices at the same time, regularization can be used to impose consistency among them.}

We aim at optimizing the energy defined in equation \ref{eq:chap3:generalSliceToVolume}, by choosing the best $\hat{\varTheta}$ that aligns the 2D and 3D images. Depending on the variables we are trying to infer, and their optimality guarantees, they can be classified in different categories. A full study of this topic is included in section \ref{sec:chap3:optimizationMethod}.

The general definition given in equation \ref{eq:chap3:generalSliceToVolume} considers a single slice as the input to the registration process. However, for the sake of completeness, in this survey we also consider the \REV{so-called} multi slice-to-volume registration approaches, for which several (but sparse) slices are registered to a full 3D volume. It is worth noting that we do not include methods that perform a prior 3D reconstruction from the input 2D slices before registration (see, for example, the work by \cite{Arbel2001}), since these methods reduce the problem to the classic 3D-3D scenario, which is not within the scope of our work. On the contrary, we consider those methods that directly register the subset of 2D slices (which can be orthogonal, parallel, arbitrary or even without an a-priori known spatial relation), rather than reconstructing a volume. Discussion and classification of the approaches according to this criterion is presented in section \ref{sec:chap3:numberOfSlices}.

\begin{landscape}
\begin{table}[]
\centering
\caption{Comparative table: It includes a selection of those papers cited in this survey that were published in the most important journals and conferences related to medical image analysis. They are classified according to the taxonomy used in this work, namely Matching Criterion (Section \ref{sec:chap3:matchingCriterion}), Transformation Model (Section \ref{sec:chap3:transformationModel}), Optimization Method (Section \ref{sec:chap3:optimizationMethod}) and Number of Slices (Section \ref{sec:chap3:numberOfSlices}).}
\label{table:comparison}
\resizebox{1.3\textheight}{!}{%
\begin{tabular}{@{}c
>{\columncolor[HTML]{EFEFEF}}c 
>{\columncolor[HTML]{EFEFEF}}c 
>{\columncolor[HTML]{EFEFEF}}c 
>{\columncolor[HTML]{EFEFEF}}c cc
>{\columncolor[HTML]{EFEFEF}}c 
>{\columncolor[HTML]{EFEFEF}}c 
>{\columncolor[HTML]{EFEFEF}}c cc@{}}
\toprule
                                     & \multicolumn{4}{c}{\cellcolor[HTML]{EFEFEF}\textbf{Matching Criterion} }       & \multicolumn{2}{c}{\textbf{Transformation Model}} & \multicolumn{3}{c}{\cellcolor[HTML]{EFEFEF}\textbf{Optimization Method}} & \multicolumn{2}{c}{\textbf{Number of slices}} \\ \cmidrule(l){2-12} 
\multirow{-2}{*}{\textbf{Reference}} & \textbf{Iconic} & \textbf{Geometric} & \textbf{Sensor Based} & \textbf{Hybrid} & \textbf{Rigid}        & \textbf{Non-Rigid}        & \textbf{Continuous}     & \textbf{Discrete}     & \textbf{Heuristic}     & \textbf{Single}        & \textbf{Multi}       \\ \midrule
\REV{\cite{Gourdon1994}}                   &                 &  X                  &                       &                & X                     &                           & X                       &                       &                        & X                      &                      \\
\cite{Kim1999}                       & X               &                    &                       &                 & X                     &                           & X                       &                       &                        & X                      &                      \\
\cite{Fei2002}                       & X               &                    &                       &                 & X                     &                           & X                       &                       &                        & X                      &                      \\
\cite{Fei2003}                       & X               &                    &                       &                 & X                     &                           & X                       &                       &                        & X                      &                      \\
\cite{Fei2004}                       & X               &                    &                       &                 & X                     &                           & X                       &                       &                        & X                      &                      \\
\cite{Fei2004a}                      & X               &                    &                       &                 & X                     &                           & X                       &                       &                        & X                      &                      \\
\cite{Park2004}                      & X               &                    &                       &                 & X                     &                           & X                       &                       &                        &                        & X                    \\
\cite{Penney2004}                    & X               &                    & X                     &                 & X                     &                           & X                       &                       &                        &                        & X                    \\
\cite{Yeo2004}                       & X               &                    &                       &                 & X                     &                           & X                       &                       &                        & X                      &                      \\
\cite{Zarow2004}                     & X               & X                  &                       & X               &                       & X                         & X                       &                       &                        &                        & X                    \\
\cite{Bao2005}                       &                 &                    & X                     &                 & X                     &                           & -                       & -                     & -                      & X                      &                      \\
\cite{Noble2005}                     & X               &                    &                       &                 & X                     &                           & X                       &                       &                        & X                      &                      \\
\cite{Rousseau2005}                  & X               &                    &                       &                 & X                     &                           & X                       &                       &                        &                        & X                    \\
\cite{Smolikova2005}                 & X               &                    &                       &                 & X                     &                           & X                       &                       &                        & X                      &                      \\
\cite{Rousseau2006}                  & X               &                    &                       &                 & X                     &                           & X                       &                       &                        &                        & X                    \\
\cite{Chandler2006}                  & X               &                    &                       &                 & X                     &                           & X                       &                       &                        &                        & X                    \\
\cite{Penney2006}                    & X               &                    & X                     &                 & X                     &                           & X                       &                       &                        &                        & X                    \\
\cite{Micu2006}                      & X               &                    &                       &                 & X                     & X                         & X                       &                       &                        & X                      &                      \\
\cite{Yeo2006}                       & X               &                    &                       &                 & X                     &                           & X                       &                       &                        & X                      &                      \\
\cite{Birkfellner2007}               & X               &                    &                       &                 & X                     &                           & X                       &                       & X                      & X                      &                      \\
\cite{Boer2007}                      &                 & X                  &                       &                 & X                     &                           &                         & X                     &                        & X                      &                      \\
\cite{Jiang2007}                     & X               &                    &                       &                 & X                     &                           & X                       &                       &                        &                        & X                    \\
\cite{Jiang2007a}                    & X               &                    &                       &                 & X                     &                           & X                       &                       &                        &                        & X                    \\
\cite{Sun2007}                       & X               &                    & X                     &                 & X                     &                           & X                       &                       &                        &                        & X                    \\
\cite{Chandler2008}                  & X               &                    &                       &                 & X                     &                           & X                       &                       &                        &                        & X                    \\
\cite{Dalvi2008}                     &                 & X                  &                       &                 & X                     &                           & X                       &                       &                        &                        & X                    \\
\cite{Brooks2008}                    & X               &                    &                       &                 &                       & X                         & X                       &                       &                        &                        & X                    \\
\cite{Gefen2008}                     & X               &                    &                       &                 &                       & X                         &                         &                       & X                      & X                      &                      \\
\cite{Hummel2008}                    & X               &                    & X                     & X               & X                     &                           & X                       &                       &                        & X                      &                      \\
\cite{Kim2008}                       & X               &                    &                       &                 & X                     &                           & X                       &                       &                        &                        & X                    \\
\cite{Kim2008a}                      & X               &                    &                       &                 & X                     &                           & X                       &                       &                        & X                      &                      \\
\cite{Leung2010}                     & X               &                    &                       &                 &                       & X                         & X                       &                       &                        &                        & X                    \\
\cite{Wein2008}                      & X               &                    & X                     & X               & X                     & X                         & X                       &                       &                        &                        & X                    \\
\cite{Wein2008a}                     & X               &                    &                       &                 &                       & X                         & X                       &                       &                        &                        & X                    \\
\cite{Xu2008}                        & X               &                    & X                     & X               & X                     &                           & X                       &                       &                        &                        & X                    \\
\cite{Yeo2008}                       & X               &                    &                       &                 & X                     &                           & X                       &                       &                        & X                      &                      \\
\cite{Fruhwald2009}                  & X               &                    &                       &                 & X                     &                           & X                       &                       &                        & X                      &                      \\
\cite{Gholipour2009}                 & X               &                    &                       &                 & X                     &                           & X                       &                       &                        &                        & X                    \\
\cite{Huang2009}                     & X               &                    & X                     & X               & X                     &                           & X                       &                       &                        &                        & X                    \\
\cite{Jiang2009}                     & X               &                    &                       &                 & X                     &                           & X                       &                       &                        &                        & X                    \\
\cite{Osechinskiy2009}               & X               &                    &                       &                 &                       & X                         & X                       &                       &                        & X                      &                     
\end{tabular}
}
\end{table}

\end{landscape}

\begin{landscape}
\begin{table}[]
\centering
\caption{Continuation of Table \ref{table:comparison}}
\label{table:comparison2}
\resizebox{1.3\textheight}{!}{%
\begin{tabular}{@{}c
>{\columncolor[HTML]{EFEFEF}}c 
>{\columncolor[HTML]{EFEFEF}}c 
>{\columncolor[HTML]{EFEFEF}}c 
>{\columncolor[HTML]{EFEFEF}}c cc
>{\columncolor[HTML]{EFEFEF}}c 
>{\columncolor[HTML]{EFEFEF}}c 
>{\columncolor[HTML]{EFEFEF}}c cc@{}}
\toprule
                                     & \multicolumn{4}{c}{\cellcolor[HTML]{EFEFEF}\textbf{Matching Criterion} }       & \multicolumn{2}{c}{\textbf{Transformation Model}} & \multicolumn{3}{c}{\cellcolor[HTML]{EFEFEF}\textbf{Optimization Method}} & \multicolumn{2}{c}{\textbf{Number of slices}} \\ \cmidrule(l){2-12} 
\multirow{-2}{*}{\textbf{Reference}} & \textbf{Iconic} & \textbf{Geometric} & \textbf{Sensor Based} & \textbf{Hybrid} & \textbf{Rigid}        & \textbf{Non-Rigid}        & \textbf{Continuous}     & \textbf{Discrete}     & \textbf{Heuristic}     & \textbf{Single}        & \textbf{Multi}       \\ \midrule
\cite{SanJoseEstepar2009}      & X &   &   &   & X &   & X &   &   & X &   \\
\cite{Elen2010}                & X &   &   &   & X &   & X &   &   &   & X \\
\cite{Gholipour2010}           & X &   &   &   & X &   & X &   &   &   & X \\
\cite{Honal2010}               & X &   &   &   &   & X & X &   &   & X &   \\
\cite{Kim2010}                 & X &   &   &   & X &   & X &   &   &   & X \\
\cite{Kim2010b}                & X &   &   &   & X &   & X &   &   &   & X \\
\cite{Tadayyon2010}            & X &   &   &   & X & X & X &   & X &   & X \\
\cite{Gholipour2011}           & X &   & X &   & X &   & X &   &   & X &   \\
\cite{Kim2011}                 & X &   &   &   & X &   & X &   &   &   & X \\
\cite{Marami2011}              & X &   &   &   &   & X & X &   &   &   & X \\
\cite{Osechinskiy2011}         & X &   &   &   & X & X & X &   &   & X &   \\
\cite{Tadayyon2011}            & X &   &   &   & X & X & X &   & X &   & X \\
\cite{Xiao2011}                & X &   &   &   &   & X & X &   &   &   & X \\
\cite{Yu2011}                  & X &   &   &   & X &   & X &   &   &   & X \\
\cite{Kuklisova-Murgasova2012} & X &   &   &   & X &   & X &   &   &   & X \\
\cite{Mitra2012}               & X & X &   &   & X &   & X &   &   & X &   \\
\cite{Yan2012}                 & X &   & X & X & X &   & X &   &   &   & X \\
\cite{Zakkaroff2012}           & X &   &   &   & X &   & X &   &   &   & X \\
\cite{Cifor2013}               & X &   &   &   & X &   &   &   & X &   & X \\
\cite{Ferrante2013}            & X &   &   &   &   & X &   & X &   & X &   \\
\cite{Lin2013}                 & X &   &   &   & X &   &   &  & X  & X &   \\
\cite{Seshamani2013}           & X &   &   &   & X &   & X &   &   &   & X \\
\cite{Su2013}                  & X &   &   &   &   & X & X &   &   & X &   \\
\REV{\cite{Chicherova2014}}    &   & X &   &   & X &   & X  &  &   & X &   \\
\cite{Eresen2014}              & X &   & X & X & X &   &   & X &   & X &   \\
\cite{Fogtmann2014}            & X &   &   &   & X &   & X &   &   &   & X \\
\cite{Fuerst2014}              & X &   & X & X & X & X & X &   &   &   & X \\
\cite{Museyko2015}             & X &   &   &   &   & X & X &   &   & X &   \\
\cite{Nir2014}                 & X & X &   &   &   & X & X &   &   &   & X \\
\cite{Rivaz2014a}              & X &   &   &   &   & X & X &   &   &   & X \\
\cite{Rivaz2014b}              & X &   &   &   &   & X & X &   &   &   & X \\
\cite{Rivaz2014c}              & X &   &   &   &   & X & X &   &   &   & X \\
\cite{Schulz2014}              &   &   & X &   & X &   & - & - & - & X &   \\
\cite{Xu2014a}                 & X &   &   &   &   & X & X &   &   &   & X \\
\cite{Xu2014b}                 & X &   &   &   & X &   & X &   &   & X &   \\
\cite{Ferrante2015}            & X &   &   &   &   & X &   & X &   & X &   \\
\cite{Ferrante2015a}           & X &   &   &   &   & X &   & X &   & X &   \\
\REV{\cite{Hallack2015}}             &   &   &   & X &   & X & X  &  &   &   & X \\
\cite{Kainz2015}               & X &   &   &   & X &   & X &   &   & X &   \\
\cite{Yavariabdi2015}          &   & X &   &   &   & X & X &   &   & X &   \\
\REV{\cite{Chen2016}}          & X &   &   &   & X & X & X &   &   &   & X \\
\REV{\cite{Guzman2016}}        & X &   &   &   & X & X & X &   &   &   & X \\
\REV{\cite{Marami2016a,Marami2016b}} & X &   &   &   & X & X & X &   &   &   & X \\
\cite{Murgasova2016}           & X &   &   &   & X &   & X &   &   &   & X \\
\cite{Porchetto2016}           & X &   &   &   & X &   &   & X &   & X &   \\
\cite{Xiao2016}                & X &   &   &   & X &   & X &   &   &   & X

\end{tabular}
}
\end{table}

\end{landscape}

\section{Matching Criterion} \label{sec:chap3:matchingCriterion}
The matching criterion (also known as (dis)similarity measure, merit function or distance function) quantifies the level of alignment between the images, and it is typically used to guide the optimization process of the transformation model. Depending on the nature of information exploited in the matching process, registration methods can be classified as iconic (we use voxel intensities to quantify similarity), geometric (we use a sparse set of salient image locations to guide the registration) or hybrid methods (we combine both strategies). In the particular case of slice-to-volume registration, there are also some approaches which instead of using image information, rely on other non-image technologies; we will refer to them as sensor based methods. 

Slice-to-volume registration excluding image based methods (i.e. sensor based slice-to-volume registration) is mainly performed using two different technologies: optical (OTS) and electromagnetic (EMTS) tracking systems \citep{Birkfellner2008}. Optical systems employ different types of cameras (such as infrared (IR), standard RGB or laser cameras) to track markers which help to identify the current position of the objects. Electromagnetic positioning systems perform tracking based on a system of transmitter, sensors and processing unit that localizes the position and orientation of a target object by measuring electromagnetic field properties. Both methods inherit constraints on the nature of transformation that they can estimate, since only rigid transformations can be obtained through these technologies. Moreover, accurate initial calibration is usually required at the beginning, due to the fact that positioning is done through the estimation of the relative displacement from the previous one and therefore errors can be propagated and accumulated through all the process. Instead, registration algorithms exploiting image based information -iconic or geometric- can deal with elastic anatomical changes and they are less sensitive to initial errors (since it is simple to correct them iteratively during the registration process). On the negative side, such methods can be more sensitive to image noise, which is frequently present in intra-operative, real-time and low-quality modalities, normally corresponding to the input 2D image. In addition, the amount of information of an image slice is sparse when compared to a volumetric image, resulting on ambiguities in terms of image matching that render the registration problem ill-posed. Different strategies have been developed to deal with these problems depending on the choice of the matching criterion.

Image registration can be monomodal (when the slice and the volume are captured with the same type of image technology) or multimodal (when slice and volume refer to different modalities, e.g. US slice and MRI or CT volume). In the former case, the task of measuring the similarity between the images is simpler, since pixel/voxel intensity values corresponding to the same anatomical structure are highly correlated, or even identical, in both images. Therefore, iconic methods may perform better since the main issue associated to them -the difficulty to explain image similarities using pixel/voxel correspondences- is already solved. In case of multimodality, where the relation between pixel intensities is not obvious, there are two major alternatives: to continue using the iconic matching criterion but defining more complex similarity measures, or to adopt a geometric or sensor based strategy which appears to be more robust when dealing with different image modalities.

\subsection{Iconic} \label{sec:chap3:iconic}
Iconic matching criteria are defined using image intensity information. Similarity between images is measured using functions that act on the pixel/voxel intensity level. Standard signal processing tools, information theoretical approaches or even similarity measures defined for particular image modalities can be considered. The challenge lies in describing both images on a common space where they can be easily compared. In particular, in the context of multimodal registration, where voxel intensities corresponding to the same anatomical or functional structures are dissimilar. There are two desirable properties sought at the definition of any iconic similarity measure: (i) to be convex, since it simplifies the optimization process and (ii) to be discriminative, in the sense that it assigns distinct values to different tissues or anatomical structures. Convexity is also associated 

Typically, in the context of slice-to-volume registration, an iconic matching criterion $\delta(I_1, I_2): \Omega_1 \times \Omega_2 \rightarrow \Re$ is defined to quantify the similarity between two slices (or patches, according to whether we specify a global or local function). Such a similarity measure varies depending on the modalities we are trying to register. In a monomodal scenario, simple similarity measures such as the sum of absolute differences (SAD) \citep{Ferrante2013, Ferrante2015, Ferrante2015a, Leung2010}, sum of square sifferences (SSD) \citep{Fogtmann2014, Heldmann2009, Leung2010, Marami2011, Miao2014, Osechinskiy2011a, Su2013, Seshamani2013, Xu2008, Yu2008, Yu2011} or even mean of square differences \citep{Fogtmann2014, Gholipour2010, Gholipour2009, Honal2010, Kim2008, Kim2010, Tadayyon2010, Tadayyon2010a} of the intensity values can be used. In vector notation, SAD can also be seen as the $L^1$ norm of vectorized image, whereas SSD would correspond to the $L^2$ norm. These metrics assume a straightforward correspondence between the intensity values in both images, which is not necessarily the case. In real clinical cases like image guided surgeris, images captured during the procedures tend to be noisy, low resolution and sometimes even different modalities have to be fused. In these cases, SSD and SAD will present poor performance.

More complex metrics, exploiting statistical properties of the observed intensity values in both images, have also been proposed. These methods, on top of handling identity transformations, can cope with piece-wise linear relationships between the intensities in the images to be registered. In these cases, image pixels in both images are seen as entries of two random vectors X and Y. Cross-correlation (CC) is a well known function widely used in the fields of signal processing and statistics, also applied in several slice-to-volume registration studies \citep{Birkfellner2007, Elen2010, Fei2002, Fei2003, Fei2003a, Fei2004, Fruhwald2009, Hummel2008, Jiang2007,Jiang2007a, Jiang2009, Kainz2015, Kim2005, Miao2014, Noble2005, Osechinskiy2009, Osechinskiy2011a, Xiao2016, Xu2008, Yan2012, Zarow2004}. CC measures the correlation between the entries of X and Y. It is simple to compute and, more importantly, invariant to shifts and scaling in the intensity domain. Another metric is the correlation ratio (CR), which has shown promising results even in multimodal image registration \citep{Roche1998}. It measures functional dependencies between X and Y, taking values between 0 (no functional dependence) and 1 (purely deterministic dependence). CR is intrinsically asymmetrical, since the two variables (images) do not play the same role in the functional relationship. In other words, unlike CC, CR offers different values depending on the order that images were considered. CR has been used as an iconic criterion for slice-to-volume registration in \citep{Marami2011, Osechinskiy2011a, Smolikova2005}.

Information theoretic similarity measures are usually the choice of preference for multimodal registration, and slice-to-volume registration is not an exception. The most popular is mutual information (MI), which measures the statistical dependence or information redundancy between the image intensities of corresponding distributions in both images, that is assumed to be maximal if the images are geometrically aligned \citep{Maes1997}. It requires an estimation of joint and marginal probability density functions (PDFs) of the intensities in every image. Given the information sparse nature of slice-to-volume registration when compared to the volume-to-volume scenario, the estimation of these joint PDFs for every slice -especially in slices of low image resolution/number of samples- is a hard task and may redound to poor MI-based registration results. An alternative approach to improve MI-based slice-to-volume registration was developed by incorporating informative PDF priors in the context of fMRI time-series registration \cite{Bhagalia2009}. First, it was shown that slices located near the middle of the head scans give more reliable PDFs and MI estimations because they refer to a richer information space than the end slices (top or bottom). End-slices registration is then guided by a joint PDF prior based on intensity counts from registered center-slices. Alternatively, a better MI calculation in slice-to-volume registration can be achieved by using a PDF estimate that retains as much information about voxel intensities as possible from the higher resolution anatomical data set, when registering 2D MR scout scan to a complete 3D MR brain volume \citep{Chandler2004}. MI is a widely used similarity measure for slice-to-volume registration, adopted in an important number of methods in the last decades, e.g. \citep{Birkfellner2007, Brooks2008, Eresen2014, Fei2002, Ferrante2015a, Fogtmann2014, Gill2008, Huang2009, Kim1999, Museyko2015, Park2004, Rousseau2006, Seshamani2013, Smolikova2005, Tadayyon2010, Xiao2011, Xu2014a, Yeo2004, Zakkaroff2012, Zarow2004, Chen2016}. One of the main drawbacks of MI is that it varies when the overlapping area between the images changes, i.e. it is not invariant to changes in the overlap region throughout registration. It could happen that while estimating the transformation model, some potential solutions lie out of the volume. In such cases, an overlap invariant function would be of choice. To this end, a modified version of MI, the normalized mutual information (NMI), can be applied, which is simply the ratio of the sum of the marginal entropies and the joint entropy \citep{Studholme1999}. Another advantage of NMI with respect to MI is its range: it conveniently takes values between 0 and 1. NMI has been used as well for slice-to-volume registration \citep{Chandler2006, Elen2010, Gefen2008, Hummel2008, Jiang2007, Kainz2015, Kuklisova-Murgasova2012, Leung2010, Marami2011, Miao2014, Rousseau2005}.

Using prior knowledge like segmentation masks during the iconic registration process can be useful \citep{Shakeri2016}. These approaches, also known as region-based methods, employ intensities information or statistics to describe a pre-segmented region. Chan-Vese metric \citep{Chan2001} for instance, aims to minimize the intensity variances on the regions inside and outside a given segmentation contour. Nir and coworkers \citep{Nir2011, Nir2014} applied this matching criterion to the problem of aligning multiple slices of histological images to in vivo MR images of the prostate. 

Border information is another low level visual cue that was exploited by iconic methods. It is usually determined from the intensity gradient of the images, which gives an idea of the image structure defined by intensity changes, independently of their actual value. MI as well as SAD or SSD can be applied on top of the gradient magnitudes of both images. In \citep{Brooks2008}, for example, MI between the gradient magnitudes of an US slice and MRI volume was used. \cite{Su2013} applied SSD on both image intensities and gradients, combining them in a unique similarity measure. In \citep{Xu2014b}, the matching criterion was defined using the normalized gradient field of the images, while CC of both intensity and gradient magnitude was adopted in \citep{Xu2008}. \REV{In \citep{Lin2013}, a merit function called Weighted Edge-Matching Score (WEMS) was used to evaluate the similarity of a 2D real-time image and the corresponding reformated slice to search for the 3D pose of a bone model.}

More robust similarity measures have been introduced more recently to deal with problems in medical image registration, and in the slice-to-volume case in particular. Remarkable contributions have been made by \citep{Wein2008, Fuerst2014} to this field. They propose different similarity measures based on the simulation of US images from MRI and from CT, which can deal with these challenging multimodal registration problems. In \citep{Wein2007, Wein2008}, novel methods for simulation of ultrasonic effects from CT data are presented, together with a new similarity measure entitled Linear Correlation of Linear Combination ($LC^2$), which is invariant to missing simulation details, yielding smooth properties and a global optimal corresponding to the correct alignment. Since they simulate US imaging effects with respect to the probe geometry, the original B-mode scan planes of the sweep are used instead of 3D reconstruction, making it suitable for multi slice-to-volume registration. $LC^2$ was used by \cite{Fuerst2014}, although without the simulation process: the similarity measure was defined by locally matching US intensities to both MRI intensity and gradient magnitude. More recently, a variation on the $LC^2$ entitled $BOXLC^2$ which produces a measurable speed improvement over $LC^2$ when running on GPU (while retaining similar performance in the context of single slice-to-volume registration) was presented in \citep{Pardasani2016}.

Another robust similarity measure, the modality independent neighborhood descriptor (MIND) was proposed by \cite{Heinrich2012} for multimodal rigid and deformable registration. It is based on the concept of local self-similarity at the patch level. They create a multi-dimensional descriptor through ranking of the local intensity distribution of the two images, therefore providing a very good representation of the local shape of an image feature. It can be computed in a dense fashion across all the pixels (or voxels) of the images; once it is computed, the SSD of the MIND representations can be used as a similarity measure. Thanks to its point-wise (pixel or voxel-wise) calculation nature, it can be adapted to almost any registration algorithm. An extension to MIND, named Self Similarity Context (SSC) is also estimated using patch-based self-similarities \citep{Heinrich2013}. In \citep{Cifor2013}, MIND was successfully used in a multi slice-to-volume registration framework to align untracked freehand 2D US sweeps to CT volumes. 

A last family of robust similarity measures was introduced by \citep{Rivaz2014a, Rivaz2014b, Rivaz2014c} in the context of mutlimodal US/MRI image registration. In \citep{Rivaz2014a}, a metric called Contextual Conditioned Mutual Information (CoCoMI) was proposed. The metric aims at tackling one of the main drawbacks of classic MI based methods, that is taking into account the intensity values of corresponding pixels and not of neighbor. Images are treated as ``bag of words'' and consequently contextual information is ignored. CoCoMI overcomes this limitation by conditioning the MI estimation on contextual information. In \citep{Rivaz2014b}, Self Similarity $\alpha$-MI (SeSaMI) -another MI based matching criterion- is proposed. $\alpha$-MI is usually calculated on multiple features like intensities and their gradients, as opposed to standard MI which is usually calculated on intensities only. SeSaMI combines this multi-feature $\alpha$-MI formulation with self-similarities in a kNN $\alpha$-MI registration framework by penalizing clusters (i.e. the nearest neighbors) that are not self-similar. Finally, in \citep{Rivaz2014c}, a CR similarity measure is introduced. Robust Patch Based Correlation \REV{Ratio} (RaPTOR) computes local CR values on small patches and adds them to form a global cost function. Authors claim a property that makes suitable such methods: their metric is invariant to important spatial intensity inhomogenity. \REV{This is especially useful} when dealing with US images due to wave attenuation, shadowing and enhancement artifacts. One of the main advantages of these Rivaz's metrics (CoCoMI, SeSaMI and RaPTOR) is that their gradient can be derived analytically, and therefore the cost function can be efficiently optimized using stochastic gradient descent methods.

\subsection{Geometric} \label{sec:chap3:geometric}
Geometric registration finds correspondences between meaningful anatomical locations or salient landmarks \citep{Joshi2000, Glocker2011}. These methods aim at minimizing an energy function that, for a given transformation, measures the discrepancy between the key-points detected in both, the 2D slice and the volumetric image. Simplicity of the registration process once the landmarks are appropriately determined, no sensitivity to initializations and a wider capture range in terms of deformation are the main strengths of such approaches. On the other hand, the landmark detection and matching processes are not that trivial, and errors on their position compromise the accuracy of the registration process. Moreover, due to the sparsity of the key-points, the quality of the deformations may become insufficient (due to the limited support on the interpolation).

The early work by \cite{Gourdon1994} presents a geometric method to perform slice-to-volume registration between a curve and a surface. In this work, Gourdon and Ayache exploited the knowledge about the differential properties computed on both, the curve and the surface, to constrain the rigid matching problem. The most relevant contribution of this work is their discussion about how differential constraints can be used to rigidly register a curve on a surface. However, they used the basic Marching Cubes algorithm \citep{Lorensen1987} to extract them from simulated and medical data. The extraction of these structures remains an arduous task for images with low resolution, a scenario often valid in slice-to-volume registration. \REV{In these cases, alternative methods (such as super-resolution reconstruction \citep{oktay2016multi} or registration-based interpolation methods \citep{penney2004registration}) could be used to improve on the quality of the original images before extracting the structures of interest, leading to more accurate results after the geometric registriation process is applied.}

Extracting distinctive features becomes even more complicated for medical images (opposed to natural ones), since the first ones are usually not as discriminative (lack of texture) as the last ones. Invariance to scaling, rotation and changes in illumination or brightness constitute useful properties for methods seeking to extract salient points. Highly distinctive features (in both spatial and frequency domains) simplify the matching task since it is likely that they are correctly matched. Classical examples of such descriptors successfully applied in different computer vision tasks are the Scale-Invariant Feature Transform (SIFT) \citep{Lowe1999}, the Harris detector \citep{Harris1988}, the Histogram of Oriented Gradients (HOG) \citep{Dalal2005}, Speeded Up Robust Features (SURF) \citep{Bay2008}, etc. More recently, features learned using deep learning have been successfully applied for scenarios involving massive amount of annotated data for training \citep{Long2014}. These features are used to extract the salient landmarks that will guide the geometric registration process. Once the landmarks are established in both slice and volumetric images, there are two choices. First, we can perform a matching process which results in establishing correspondences between pairs of points from the two sets. Once the matching process is finished, correspondences can be used to estimate the desired -rigid or non-rigid- transformation models. Alternatively, methods directly estimating the transformation model without inferring any correspondence could also be applied. Sometimes, both correspondences and transformation model can be inferred at the same time, like in case of Iterative Closest Point (ICP) \citep{Besl1992} algorithm. ICP is an algorithm subsequently improving the matching of point pairs. It minimizes the sum of geometric distances between the transformed set of source image landmarks and the closest detected landmarks in the target one. It is a simple and fast method that follows the closest neighbor principle to perform the matching task, which often converges to local minimum though.

\cite{Dalvi2008} proposed a slice-to-volume registration approach which uses ICP. First, it extracts phase congruency information from the slices/volume using oriented 2D Gabor wavelets. Then, using local non maximum suppression, a robust and accurate set of feature points is automatically obtained, which are subsequently matched by ICP inferring a rigid body transformation. \cite{Nir2014} presented a particle filtering framework using a geometric matching criteria where the closest point principle is applied (they use the SSD between pairs of closest points in the two sets). In their proposal, the inferred transformation model involves 9-DOF (affine) matrices instead of 6-DOF transformations (rigid body). Nir's method was applied by \cite{Moradi2013} as a pre-processing step in a tumor detection learning based framework, to register US to whole-mount histopathology references of prostate images, mapping the location of tumors to the US image domain. Another method that uses ICP was proposed by \cite{Yavariabdi2015} for deformable registration. They register 2D transvaginal US (TVUS) images with 3D MRI volumes to localize endometrial implants. They use contour to surface correspondences through a novel variational one-step deformable ICP method, finding a smooth deformation field while establishing point correspondences automatically. The main drawback of this approach is that it relies on the user to segment the organs. Manual pelvic organ segmentation is a laborious and time consuming task, and consequently the applicability of the approach in real scenarios is limited. Note that in the papers discussed in this paragraph, different types of transformation models where estimated based on the closest point principle, showing the flexibility of such approach. 

Signed distance functions (or signed distance maps) can be used to avoid the landmark matching step when performing non-rigid registration of shapes and points. In this case, the landmarks or shapes are assigned to zero distance, while the rest of the pixels of the image are labeled with the distance to the nearest geometrical primitive (landmark or boundary). Once the distance map is created, the optimal transformation model can be estimated by means of standard iconic registration (e.g. using SSD on the distance maps)\citep{Paragios2003, Huang2006, Taron2009}. In \citep{Boer2007}, signed distance functions are used to register slices of histological images with a pre-reconstructed 3D model. Authors proposed to use a naive brute force approach for optimization, which incurs in extremely high computational cost.

\subsection{Sensor Based Methods} \label{sec:chap3:sensorBased}
Sensor based systems are an alternative to image-based methods (even if images are often used for their initialization), where slice location and orientation are continuously updated using sensors. Such information can be used to extract the corresponding slice from a given volume assuming an initial correspondence. Optical (OTS) and electromagnetic (ETMS) tracking systems are the most widely used technologies to perform this task. OTS systems determine in real-time the position of an object by tracking the positions of either active or passive markers attached to the object. It requires a line-of-sight to be maintained between the tracking device and the instrument to be tracked. This might cause an inconvenience to the physician -especially in image guided interventions- during his work, resulting highly inconvenient. EMTS systems usually consist of three basic components: the electromagnetic field generator, a system control unit that interfaces with a PC, and tracked sensor coils together with their interfaces to the system control unit. By measuring the behavior of each coil, the position and orientation of the object can be determined. EMTS do not inherit the same constraints as the line-of-sight requirements but it is sensitive to distortion from nearby metal sources and exhibits limited accuracy compared to optical tracking \citep{Birkfellner2008}. Moreover, neither OTS nor EMTS can deal with elastic deformations between volume and slice. In general, one can claim that sensor based methods are more reliable (when they are calibrated correctly) than image based methods since they are not affected by ambiguities that arise during the image interpretation process. 

In \citep{Gholipour2011}, a three-dimensional magnetic field sensor is used to track the motion of a subject during MRI scanning. It allows estimating the location of the slices for volume reconstruction. The rigid 6-DOF transformation of the sensor in three dimensions (6-DOF) is calculated in real-time using the native gradient fields of the MRI scanner. The relative 3D location of each slice is computed through the sensor motion parameters at the time of slice acquisition. Other works exploiting EMTS systems can be found here \citep{Hummel2008, Olesch2011, Olesch2011a, Xu2008}.

Optical tracking systems can also replace classical image based slice-to-volume registration algorithms. In \citep{Schulz2014} a slice-by-slice motion correction for fMRI image reconstruction (see section \ref{sec:chap3:motionCorrectionVolumeReconstraction} for a complete discussion about this problem) is achieved thanks to an optical tracking system. Authors proposed to use three tracking cameras with embedded image processing, to track the position of the three optical markers attached to the skull using goggles. Tracking information is then used to replace the classical slice-to-volume registration step \citep{Kim1999} necessary to account for motion correction during fMRI image reconstruction. In \citep{Bao2005}, optical tracking is used to register laparoscopic US to CT images of a phantom liver, based on an infrared camera. The advantage of infrared cameras is that the position sensor does not pick up interference from reflections and ambient light. However, the line-of-sight requirement still holds and therefore sensor must not be blocked. Other papers using OTS to track the position of image slices when registering them with volumetric images can be found here \citep{Huang2009, Penney2006, Penney2004, Schulz2014}. 

More recently \citep{Eresen2014}, smartphone tracking technology was considered as a navigation tool to initialize a slice-to-volume registration process between a histological 2D slice and a MR volume. The Inertial Measurement Unit (IMU) of the smartphone (also available as a standalone component) is used to define the orientation of the slice. Given the orientation and position data of an IMU sensor, one can then interpolate the corresponding slice from the MR volume. This interactive alignment is applied to determine initial orientation for the 2D slice, and is refined using an iconic MI based registration process optimized via brute force. IMU tracking systems use a combination of accelerometers and gyroscopes to measure acceleration and angular velocity, respectively. Since acceleration is the second derivative of position with respect to time, and angular velocity is the first derivative, angular changes integrated over time from a known starting position yield translation and rotation (i.e. a 6-DOF transformation) \citep{Birkfellner2008}. These sensors are cheap and widely available. However, a major issue of using IMUs for tracking refers to accumulation of errors (either systematic or statistical), leading to degradation estimation over time. Kalman filters \citep{Kalman1960} can be used to deal with this type of issues and have been often adopted from the scientific community. \REV{In \citet{Marami2016a, Marami2016b} for example, Kalman filters were coupled with slice-to-volume registration to estimate head motion parameters.}

Note that, as described in the next section, several papers that propose sensor based methods, are actually a combination of this technology with some image-based registration technique, resulting in what we call a hybrid method.

\subsection{Hybrid} \label{sec:chap3:hybrid}
\cite{Sotiras2010} states that the hybrid matching criteria take advantage of both, iconic and geometric approaches, in an effort to get the best of both worlds. This category can be extended by including sensor based technology as it can also be combined with iconic or geometric methods. \REV{In this case, sensors are used to initiallize the registration process or to provide a real-time update for the slice position which can then be refined using image-based registration.}

\cite{Mitra2012} proposed a hybrid slice-to-volume registration approach, combining geometric and iconic matching criteria in a probabilistic framework to register transrectal US (TRUS) with MR images of the prostate. The geometric component is based on shape-context \citep{Belongie2002} representations of the segmented prostate contours. Bhattacharyya distance \citep{Bhattachayya1943} between the shape-context histograms of the two shapes is used to find point to point correspondence in every axial MR image. The Chi-square distances between the TRUS slice and each of the MR slices are calculated and used to determine the matching slice. Once the TRUS-MR slice pair with the minimum Chi-square distance is determined, it is used to retrieve a 2D rigid transformation (in-plane rotation and translation) between them. This transformation is applied to the remaining MR slices to ensure similar 2D in-plane rigid alignment with the 2D TRUS slice. The iconic step is performed by measuring the similarity (NMI and CC are used as metric) between the TRUS image and every rotated axial slice of the MRI. Finally, shape and image similarity measures are transformed into probability density functions and mapped into a statistical similartity framework towards retrieving the MR slice that resembles the TRUS image. Two issues limit the applicability of such an approach in clinical practice. First, the registration is determined by the manual segmentations of the prostate on both images. Second, the method assumes that the TRUS slices are parallel to the MR axial planes; therefore, the result of the registration does not consider any out-of-plane rotation, which could occur in a realistic scenario.

Tracking information (coming from a EMTS or OTS system) can be combined in a hybrid approach with iconic or geometric criteria to perform slice-to-volume registration. \REV{In this case, the tracking signal could be used to intialize the process or calculate the additive position of a new 2D slice with respect to the previous one, bringing them to the same coordinate system. Iconic or geometric registration can then be incorporated to complement this process. In \cite{Penney2004} for example, image-based registration was used to calculate the initial transformation that relates this unique coordinate system with the volumetric image, and sensor-based updates for the relative position of the slice where then used to perform the tracking. This approach was used by \cite{Penney2004} to register intra-operative US images to a pre-operative MR volume. The algorithm was extended in \citep{Penney2006} to deal with 2D US to volumetric CT registration.} \cite{SanJoseEstepar2009} refines the rigid tracking information provided by an EM system attached to a laparoscopic or endoscopic probe. The refinement is made in terms of translation parameters, using an edge-based iconic registration method combined with a phase correlation technique. In \citep{Xu2008}, EM tracking and intra-operative iconic image registration are used to superimpose MRI data on TRUS images of the prostate. To this end, a three-step algorithm is applied, with the intermediate one corresponding to slice-to-volume registration. At the beginning of the TRUS procedure, the operator performs a 2D tracked axial sweep from the prostate's base to its apex, which is used to produce a volumetric US reconstruction of the prostate volume. The reconstructed US volume and the MRI are manually aligned. Thus, 2D TRUS slices are registered to the US reconstructed volume (reducing the MRI/US multimodality problem to US/US monomodality) while the estimated transformation can still be used to recover the corresponding MRI slice. During the intervention, out-of-plane motion compensation is achieved using intermittent multi slice-to-volume registrations between nearly real-time 2D US images and the 3D US image. This rigid body registration is based on minimizing SSD with a Simplex algorithm. The final step refers to a 2D-2D registration of US (using CC and gradient based similarity measures) and it seeks to account for in-plane misalignment. In this workflow, EM tracking information is used to simplify the problem from multi to monomodality registration, and combined with iconic slice-to-volume registration to account for motion compensation. Other hybrid methods that combine tracking information with iconic or geometric registration approaches are \citep{Sun2007, Yan2012, Xiao2016}.	 

In \citep{Ghanavati2010} a method that combines the information from a tracked freehand 2D US transducer with an iconic matching criterion based on US image simulation was proposed, in the context of Total Hip Replacement (THR) surgeries. They use a statistical shape model (SSM) of the pelvis, constructed from several CT images. They developed a multi-slice to volume registration method, to register multiple 2D US slices to a statistical atlas of the pelvis. The mean shape of the atlas is first registered rigidly to the freehand 2D US images, based on an US simulation method \citep{Wein2008}. Then, the atlas is instantiated by measuring the iconic similarity between the actual US and the US images simulated from the instance of the atlas. This two-step algorithm estimating the rigid transformation and followed by the optimization of the deformable parameters, was turned into a one-shot optimized approach in \citep{Ghanavati2011}. 

In \citep{Wein2008a}, a position sensor attached to the patient's skin was used to extract a scalar surrogate measurement, which represents an anterior-posterior translation used to detect and compensate for respiratory movements of the liver. This information is combined with an iconic criterion to estimate an affine 12-DOF transformation model which maps pre-operative plans and imaging into the interventional scenario. 

\cite{Huang2009} proposed a hybrid slice-to-volume registration strategy exploiting multiple temporal frames. A multimodality image navigation system was introduced that integrates 2D US images with pre-operative cardiac CT volumes, using electrocardigram (ECG) information, optical tracking and iconic MI based registration. The ECG is used to synchronize the US images to the corresponding pre-operative dynamic 3D CT, depending on the cardiac phase indicated by the ECG. Spatial information given by the optical tracker is used to produce a near-optimal starting point for every slice, that is refined by maximizing the MI similarity measure. Such an approach works in real-time, making it applicable to real interventional scenarios. A different combination of sensors to solve the registration problem was proposed in \citep{Hummel2008}, where EM and OTS sensors are combined with iconic registration to register endoscopic US to CT data. 

\section{Transformation Model} \label{sec:chap3:transformationModel}
\begin{figure}[t!]
  \centering
     \includegraphics[width=\textwidth]{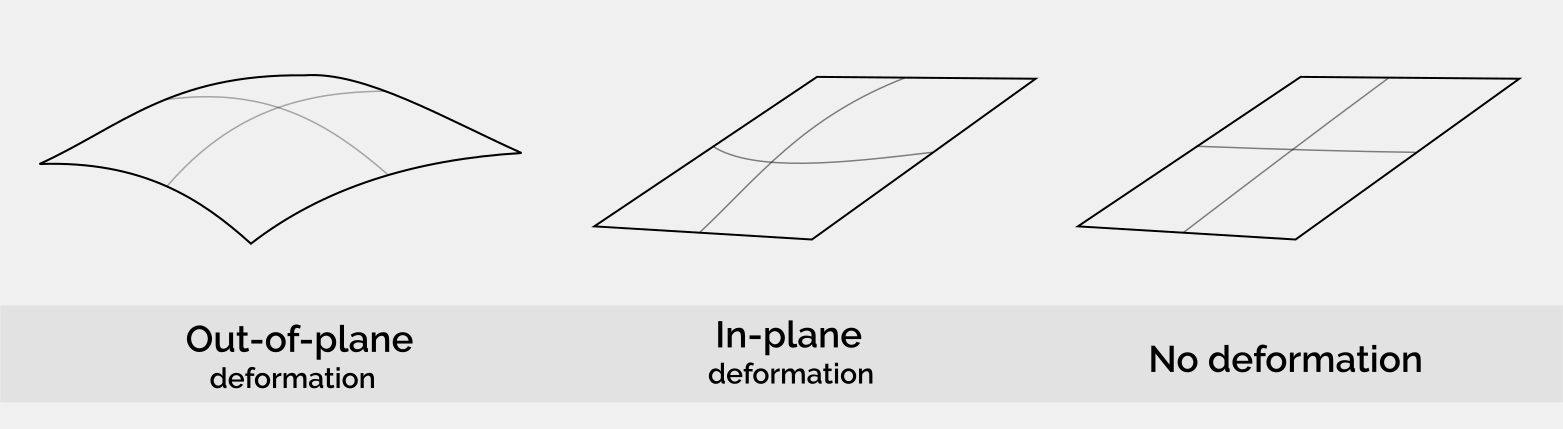}
     \caption{\REV{Depending on the application, the slice-to-volume registration algorithm may allow for out-of-plane, in-plane or null deformations. }}
  \label{fig:deformations}
\end{figure}

Transformation models explain the relation between the slice and the volume being registered, and are the outcome of the registration process. They are often classified according to their degrees of freedom. Rigid transformations deal with global rotations and translations, while deformable models -the complex case- can produce local in-plane and out-of-plane deformations \REV{(see Figure \ref{fig:deformations})}. The richness of the deformation model is proportional to the number of parameters we need to specify it, and therefore the trade-off is to be found between the model complexity and power of expression. 

\REV{When a transformation is applied to a given image (at the final step of the registration process or during the intermediate optimization steps), an interpolation method must be used to infer the intensity value of every pixel in the space of the transformed image during resampling. In this regard, there are two alternatives to consider when it comes to slice-to-volume registration: to resample the slice or the volume using the estimated transformation. The most commonly adopted strategy consists in resampling the volumetric image, since resampling a slice in 3D is not as well defined as it is for a volume. However, in some special cases where the transformation model only allows in-plane deformations in the space of the 2D image (like \citep{Ferrante2013, Ferrante2015, Ferrante2015a}), slice resampling can be performed.}

\subsection{Rigid} \label{sec:chap3:rigid}
\begin{figure}[t!]
  \centering
     \includegraphics[width=\textwidth]{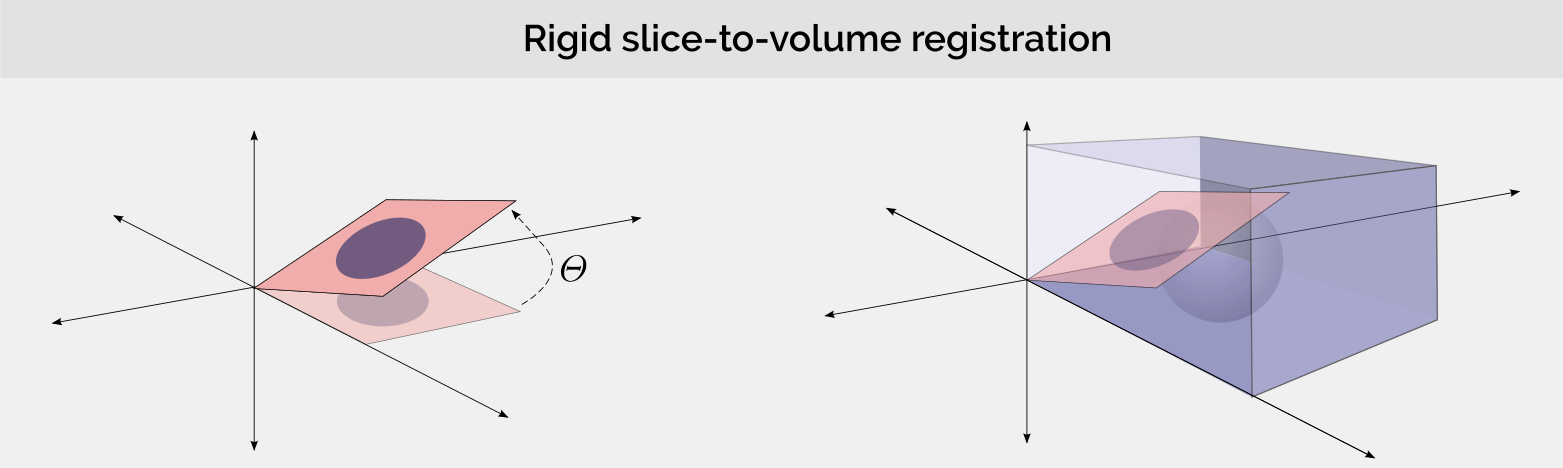}
     \caption{\REV{Rigid slice-to-volume registration. In this case, the transformation $\varTheta$ parameterizes a simple 6-DOF transformation which can only model rotation and translation. Even if simple, this type of transformation has been largely adopted by the works reviewed in this survey (see Table \ref{table:comparison}) }}
  \label{fig:slicetovolume}
\end{figure}

The simplest transformation model accounts for rotation and translation parameters. It is usually expressed as a 6 degrees of freedom (6-DOF) transformation $\varTheta^R$ composed by 3 rotation and 3 translation parameters. Such a basic model is the most common choice in the literature for slice-to-volume registration. Rigid transformations are expressive enough to explain simple slice-to-volume relations \REV{(see Figure \ref{fig:slicetovolume})}. They can deal with in-plane and out-of-plane translations and rotations. Clinical scenarios that do not inherit image distortion -like simple inter-slice motion correction \citep{Jiang2007a, Kim2008, Rousseau2005, Smolikova2005} or basic nature image guided surgeries \citep{SanJoseEstepar2009, Birkfellner2007, Gill2008}- can be modeled with rigid transformations. 

When out-of-plane motion is avoided, even simpler models can be used. \cite{Zakkaroff2012} proposed to recover in-plane slice rotations in cardiac MR series, using the stack alignment transform. In-plane translation along X and Y and rotation around a user-supplied center of rotation for the individual slices were parameterized independently. It also includes a parameter for global translation along the Z direction. Hence, it combines the individual slice transforms into a unique space of parameters -that contains 3-DOF per slice instead of 6- so that all of them can be optimized simultaneously. This type of models has been largely criticized suggesting that (especially in the context of volume reconstruction) the exact corresponding slice can only be found through a slice-to-volume registration which considers out-of-plane rotation and translations as well. In \citep{Xiao2011}, authors considered a slice-to-slice matching initial step, where every 2D image from a histological volume is matched to one axial MRI slice, via a group-wise MI based comparison. It implies a restricted transformation model involving simple slice correspondences (1-DOF). Subsequently, they correct the out-of-plane misalignment applying 2D-2D affine registration for each pair of matched slices, and a posterior 3D-3D affine registration between the histology pseudo-volume (reconstructed after transforming every 2D slice) and the MRI volume. Restricted rigid body transformations can be a convenient initialization component of a complete slice-to-volume registration pipeline.

6-DOF rigid body transformations are part of nearly all slice-to-volume registration algorithms. Literature seeking deformable registration, often initially employs rigid alignment to account for big range displacements before performing local deformable mapping (examples where this two-steps approach is applied can be found in \citep{Tadayyon2011, Xu2014a, Fuerst2014,Hallack2015,Murgasova2016}). The standard way to estimate 6-DOF rigid transformations, consists in minimizing an energy functional (based on an iconic or geometric matching criterion) often with a continuous optimization algorithm (see section \ref{sec:chap3:continuous}) where the search space is part of the Euclidean group $SE(3)$ of rigid transformations. Sensors as described in section \ref{sec:chap3:sensorBased} can be used as well for rigid alignment as proposed in a number of papers~\citep{Chandler2008, Elen2010, Fei2002, Fei2003, Fruhwald2009, Fuerst2014, Gholipour2010, Huang2009, Jiang2007, Kainz2015, Kim1999, Kim2010, Nir2011, Penney2004, Rousseau2006, Smolikova2005, Xu2014b, Yeo2004, Yu2011, Chen2016}.

\subsection{Non-Rigid} \label{sec:chap3:non-rigid}
In this review, we call non-rigid transformations to those models which perform -at some extent- changes in the structure of the images. These changes vary from simple operations -which can still be modeled using linear transformations, such as scale, flip or shear- to more complex models that produce local deformations.

\subsubsection{Linear Models} \label{sec:chap3:linear}
Linear transformations are the first step towards non-rigid registration. In \citep{Wein2008}, authors suggested that affine models can eliminate most of the large-scale deformations when registering intra-operative 2D US slices to CT images. These deformations are particularly observed between CT and US exams in different respiratory configurations, and could be handled with affine transformations. However, in a different scenario where histological slices of the prostate are registered to a MRI volume, Nir and coworkers \cite{Nir2014} concluded that the affine transformation model was unable to capture deformations, especially in the rectum area. This is due to the fact that local deformations cannot be expressed from global affine models. In such cases, deformable registration is required (further described in the next section). 

Recently, affine models were used for registering 2D slices and 3D images of the bones \citep{Hoerth2015}. Their proposal includes a novel way to initiate the registration between 2D slices and 3D images of the bones. The method uses the Generalized Hough Transform (GHT) \citep{Ballard1981} to identify suitable starting positions. They create a template version of the 2D slice by thresholding its gradient vector field, used to explore the 3D volume trough a GHT shape-detection process, providing a set of initial positions. Standard affine registration process is then used to deform images according to the initial configurations of the GHT space and updated using a MI based criterion. Such a principle could be considered to handle alternative transformation models. Other papers that estimate affine models to solve slice-to-volume registration are \citep{Gefen2008, Micu2006, Museyko2015, Wein2008a, Xiao2011, Xiao2016}. 

\subsubsection{Deformable Models} \label{sec:chap3:deformable}
Elastic deformations are powerful transformations with strong adoption in slice-to-volume mapping. The expressive power of these models depends on the number of parameters used to define them. An extensive description of different types and categories of deformable models used for non-rigid registration is presented in \citep{Sotiras2013} and \citep{Holden2008}. Here, we focus on models applied for slice-to-volume mapping.

Thin-Plate Splines (TPS) are frequently used to generate a dense deformation field from a sparse set of control points. These methods involve a set of control points that can be located in arbitrary positions, which are usually obtained by detecting salient structures. Radial basis function (RBF) -where the value at any interpolation point is given as a function of its distance from the control points- are centered at the control points and combined with affine terms to define an interpolation function. TPS minimize a bending energy based on this interpolation function, which gives a closed-form solution whose uniqueness is guaranteed in most cases. TPS can be decomposed into an affine and a local component. In \citep{Osechinskiy2011, Osechinskiy2011a}, TPS are used to parametrize a smooth 3D deformation of a 2D surface (slice). Control points are placed in a regular grid on the 2D image domain, and a 3D warp is defined using three independent TPS functions. Similarly, in \citep{Miao2014}, several regular 2D grids of control points define one TPS deformation model per slice in a multi slice-to-volume registration scenario. \cite{Kim2000} also applied TPS to a single slice-to-volume registration problem, and compared two variants of the elastic mapping (involving different number of control points to support the TPS model) with a rigid body registration algorithm. Authors extracted two main conclusions: (i) TPS based registration outperformed rigid body registration, at least in their multi slice fMRI scenario where local deformations are encountered due to the local field induced deformations or localized out-of-plane motion artifacts; and (ii) the number and location of the control points have a significant impact on the final results. 

Free-form deformation (FFD) models are also widely applied for medical image registration. Originated in computer graphics \citep{Sederberg1986}, they became popular thanks to the seminal work by \cite{Rueckert1999}. In this model, the weighting functions are cubic B-splines. The control points, have limited local support (in contrast to TPS where the control points influence the complete domain during interpolation) and are uniformly distributed over the image domain in a grid-like manner \citep{Glocker2011}. In \citep{Ferrante2013, Ferrante2015, Ferrante2015a}, FFDs were used in a graph-based discrete optimization framework to perform slice-to-volume deformable mapping. In this model, a 2D grid-like graph encodes at the same time the plane position (rigid body transformation) and the in-plane deformation of a slice with respect to its corresponding position at the 3D. The in-plane deformation is obtained through a FFD interpolation, where the nodes of the graph play the role of control points. Similarly, but in a continuous setting, \cite{Osechinskiy2011a} optimizes the position of the control points defined on a 2D grid, with unknown variables corresponding to 3D displacements. In a different scenario, \cite{Fuerst2014} proposed the use of a 3D grid based FFD within the bounding box of a tracked US sweep. Therefore, they estimated a single 3D deformation field out of a 3D grid, which can be used to deform all the slices contained in the US sweep. The main advantage of the FFD over TPS, is that it does not require solving a linear system for computing weights from displacements. However, FFD imposes constraints in terms of modeling, since the control points must be placed in a regular grid and must enclose the domain boundaries, whereas for TPS, control points can -in principle- be arbitrarily placed in the domain \citep{Osechinskiy2011a}. Other interesting works combining B-splines to model a deformable transformation in the context of slice-to-volume registration can be found here \citep{Brooks2008, Honal2010, Museyko2015, Su2013, Rivaz2014a, Rivaz2014b, Rivaz2014c, Tadayyon2010, Tadayyon2011, Xu2014a}.

An alternative model based on finite-element (FE) meshes has been used in \citep{Marami2011} to model a dynamic linear elastic deformation field. It imposes a regularization constraint on permissible volume transformations based on an iconic image similarity criterion. The advantage of FE models is that they allow incorporating the dynamic behavior of tissue deformation into the registration process, through physically meaningful constraints. 

\section{Optimization Method} \label{sec:chap3:optimizationMethod}
Optimization methods aim to determine the instance of the transformation model that minimizes a function based on the aforementioned matching criterion (see section \ref{sec:chap3:matchingCriterion}). Depending on the nature of the variables being involved, those methods can be classified to continuous or discrete. The continuous approaches exploit the entire space of parameters, while the discrete ones a discretized/quantized version of the admissible solutions.

Numerous problems in computer vision and medical imaging are inherently discrete (like semantic segmentation); however, this is not the case of slice-to-volume image registration, where the search space is continuous. Most of the published methods about slice-to-volume registration adopt a continuous formulation. Nevertheless, recent works on image registration have focused on discrete formulations \citep{Glocker2011, Glocker2008}, both for projective and sliced 2D/3D registration \citep{Zikic2010, Zikic2010a, Ferrante2013, Ferrante2015, Ferrante2015a}. The next subsections present existing work and discuss the limitations of continuous and discrete inference methods in the context of slice-to-volume registration. We also consider a third category of heuristic methods, which are independent of the nature of the variables and can be applied to a wider range of problems at the expense of not providing optimality warranty. These methods are usually applied when finding an optimal solution is impossible or impractical given the nature of the problem objective function.

\subsection{Continuous} \label{sec:chap3:continuous}

Continuous optimization algorithms are usually iterative methods. They infer the best value for a set of parameters by iteratively updating them. A common mathematical formulation for this strategy  is given by: 

\begin{equation}
 \varTheta_{t+1} = \varTheta_t + \omega \mathbf{d_t} \hspace{15px} t=0,1,2,3 ....
 \label{eq:chap3:continuousOptimization}
\end{equation}
where $\varTheta$ is the vector of parameters, $\mathbf{d_t}$ is the search direction at iteration $t$ and $\omega$ is the step size or gain factor. The search direction can be calculated using different strategies.

Continuous optimization methods can be classified according to different criteria (e.g. convex or concave minimization, solvers for linear or non-linear functions, smooth or non smooth problems, etc.). Here we simply classify them depending on whether they perform gradient- or non-gradient-based optimization, i.e., whether they exploit first (or higher order) derivatives of the energy function to compute the search direction, or they rely on other strategies. An interesting comparative analysis where methods coming from both classes are used to solve slice-to-volume registration of histological slices with MR images is presented by \cite{Osechinskiy2011a}. 

\subsubsection{Gradient based Methods}  \label{sec:chap3:continuousGradientBased}
Gradient-driven methods use the derivative of the objective function to guide the optimization process. In case of convex and differentiable functions, these methods are endowed with optimality guarantees. Otherwise, convergence to local minima is possible, and they could be sensible to initialization. Their main drawback is the requirement of analytical derivation or numerical estimation of the energy function derivatives, reducing their applicability since it is usually complicated.

Gradient descent is the simplest strategy in this category, where the search direction $\mathbf{d_t}$ is given by the negative gradient of the energy function. It refers to the standard continuous optimization method, and it has been widely applied to the problem of slice-to-volume registration \citep{Rousseau2005, Rousseau2006, Bhagalia2009, Huang2009, Tadayyon2010, Tadayyon2011, Kim2010a, Marami2011, Yu2011, Zakkaroff2012, Miao2014, Xu2014a}.  

Conjugate gradient methods use conjugate directions instead of the local gradient to estimate $\mathbf{d_t}$. Energy function with the shape of a long and narrow valley, can be optimized using fewer steps than standard gradient descent approach, resulting in faster convergence. \citep{Elen2010, Osechinskiy2011a} have applied this strategy to estimate rigid and non-rigid slice-to-volume mapping functions, respectively.

Quasi-Newton optimization strategies are used when the Jacobian or Hessian of the energy function cannot be calculated or it is too expensive to be computed. In this case, the search direction at time $t$ is computed based on an estimation of the Hessian, calculated using information provided by the previous iteration $t-1$. Quasi-Newton optimization was applied to perform deformable registration of brain slices to MR images in \citep{Kim2005}, using several similarity measures. It was also used to register 2D intra-operative US images with pre-operative volumes in \citep{Brooks2008}. One of the most popular quasi-Newton methods is the Broyden-Fletcher-Goldfarb-Shanno (BFGS) algorithm used in \citep{Honal2010} to correct for breathing motion artifacts during MRI acquisitions. A limited-memory version of BFGS (L-BFGS) is particularly suited for problems involving large numbers of variables and was applied to solve multi-slice to volume registration in \citep{Fogtmann2014, Xu2014a}.

Gauss-Newton methods can deal with non-linear least squares functions. An approximation of the Hessian matrix is used which, once combined with the gradient, provides a good estimate of the search direction for such functions. Thus, it only requires computing first order derivatives (in contrast to the standard Newton method where the actual Hessian matrix must be computed). In \citep{Heldmann2010, Olesch2011, Olesch2011a}, Gauss-Newton method is used for multimodal registration, where 2D US slices of the liver were registered to a pre-operative CT volume. Their matching criterion computes the Sum of Squared Differences (SSD) on the segmentation of the liver vessel structure in both, 3D and 2D slices. SSD is a least square function and therefore, Gauss-Newton method can be efficiently applied. This method was also used in \citep{Lasowski2008}, solving an Iteratively Reweighter Least Squares (IRLS) process in order to estimate a rigid transformation between fluoroCT and CT images. The Levenberg-Marquardt algorithm is an alternative approach to solve non-linear least squares problems, that requires also only first order derivatives. For well-behaved smooth functions, this algorithm can take more time than standard Gauss-Newton method. However, the method is more robust than the standard Gauss-Newton, meaning that it could find a solution even with bad initializations. To the best of our knowledge, this optimization algorithm was only applied to estimate rigid body transformations; however, it has been used to optimize several iconic matching criteria in a variety of domains such as reconstruction of 3D cell images \citep{Yu2011}, endoscopic interventions \citep{Hummel2008} and fetal brain imaging \citep{Kim2008, Kim2010}.

\subsubsection{Non-Gradient based Methods} \label{sec:chap3:continuousNonGradientBased}
Derivative free methods eliminate the differentiability condition of the objective function. They are applicable to a wider range of functions, including noisy, non-differentiable or even unknown functions, where we only have a black-box that returns an output value given a set of input parameters. These are cases that frequently arise when dealing with medical image registration.

The simplest non-gradient based method that has been used to solve slice-to-volume registration is the local search or best neighbor method. In this case, each parameter to be estimated is perturbed in turn using the stepsize $\omega$, and the value of the objective function is calculated. The search direction $d_t$ is then estimated as the one that produced the greatest improvement in the objective function. The main drawback of this approach is its dependency from the choice of the initial stepsize $\omega$. A big stepsize can result in an algorithm moving outside its capture range. On the other hand, for stepsizes that are too small, the optimization may become trapped in a local optimum \citep{Penney2004}. Such a greedy algorithm has been applied to estimate rigid body and affine transformations mapping slice-to-volume, which have less DOF than deformable models. Different image modalities like US to MRI \citep{Penney2004} or CT \citep{Penney2006, Sun2007} images, as well as fluoroCT images \citep{Micu2006} were considered in the clinical setting.

Nelder-Mead \citep{Nelder1965} (also known as downhill simplex method or amoeba method) is the most popular derivative-free method. It relies on the notion of simplex (a $n+1$ polytope living in a $n$-dimensional space) to explore the space of solutions in a systematic way. At every iteration, the method constructs a simplex over the search surface, and the objective function is evaluated on its vertices. The algorithm moves across the surface by replacing, at every iteration, the worst vertex of the current set by a point reflected through the centroid of the remaining $n$ points. The method can converge to a local optimum for objective functions that are smooth and unimodal. However, it exhibits more robust behavior for complex parameter space compared to standard gradient-based methods, providing a good compromise between robustness and convergence time \citep{Leung2010}. It has been widely used for slice-to-volume registration applications \citep{Chen2016, Fei2002, Fei2003, Fei2003a, Fei2004, Fruhwald2009, Gill2008, Hummel2008, Kim1999, Kim2000, Leung2010, Noble2005, Osechinskiy2011a, Park2004, Wein2008, Xiao2011, Xu2014b, Yeo2006, Yeo2004, Xiao2016}.

Another popular non-gradient based method is the well-known Powell's method. It explores the search space by performing bi-directional searches along N different vectors. Usually, these vectors initially refer to the canoninc directions. Then, the search directions are updated using linear combinations of the earlier ones. The algorithm iterates until no significant improvement is made. \cite{Smolikova2005} applied Powell's method to register two dimensional cardiac images to preoperative 3D images. Authors claim that similarly to all local techniques, Powell's method converge to local minima, but it is generally robust and accurate, and exhibits fast convergence. \citep{Wein2008a} takes advantage of the fact that Powell's method performs line search in specific directions. They apply Principal Component Analysis (PCA) on the 12 parameters of an affine transformation, reducing the search space to the 3 most significant PCA modes. Therefore, Powell direction search is initilized with the most significant PCA modes, which assures good performance and robustness. Other papers applying Powell's method to our problem can be found here \citep{Fei2003, Gholipour2010, Gholipour2009, Jiang2007, Jiang2007a, Jiang2009, Micu2006, Osechinskiy2011a}.


\subsection{Discrete} \label{sec:chap3:discrete}

In a discrete scenario, slice-to-volume registration can be expressed as a discrete labeling problem over a Markov Random Field (MRF) \citep{Wang2013}. Discrete optimization of MRFs is, in general, an NP-hard problem \citep{Shimony1994}. However, in special cases, it can benefit from very efficient solutions. The trivial brute force algorithm (i.e. trying all possible combination of labels for each and every variable) has an exponential complexity that makes such an approach unsuitable. More efficient algorithms have been developed during the last two decades which boosted the use of graphical models in the field of computer vision. They can be classified in three main categories according to \cite{Kappes2013}:

\begin{enumerate}
 \item Polyhedral and combinatorial methods, solving a continuous linear programming (LP) relaxation of the discrete energy minimization problem. The central idea is to relax the integrality condition of the variables in order to simplify the problem. Once the integrality constraint is relaxed, standard linear programming methods can be applied to solve the optimization problem, and rounding strategies are used to recover the integral solution. Examples of such approaches are Linear Programming Relaxations over the Local Polytope, Quadratic Pseudo Boolean Optimization (QPBO) \citep{Rother2007} and Dual Decomposition \citep{Komodakis2011}.
 
 \item Message passing methods, in which messages are calculated and propagated between nodes in a graph. This propagation can be seen as a re-parametrization of the original problem aiming to establish special properties in the re-weighted function that makes inference easier. Examples are the standard Loopy Belief Propagation (LBP) \citep{Murphy1999} and Three Re-weighted Belief Propagation (TRBP) \citep{Wainwright2005}.
 
 \item Max-flow and move-making algorithms make use of the well \REV{known} max-flow min-cut \citep{Boykov2004} algorithm from graph theory, which can optimally solve some instances of discrete energies. These methods are usually combined with greedy strategies that iteratively minimize over the label space by solving a sequence of max-flow min-cut sub problems. Examples are $\alpha$-expansion, $\alpha\beta$ swap \citep{Boykov2001} and FastPD \citep{Komodakis2007} algorithms. Simpler move-making algorithms not using max-flow, but still applying the strategy of starting with an initial labeling and iteratively moving to a better one until a convergence criterion is met, are the classical Iterated conditional modes (ICM) \citep{Besag1986} and its generalization Lazy Flipper\citep{Andres2012}.
\end{enumerate}

Graphical models and discrete optimization are powerful formalisms that have been successfully used during the past years in the field of computer vision \citep{Wang2013}. \REV{In general, a graphical model is represented as a graph $\mathcal{G = \langle V, E \rangle}$ where vertices in $\mathcal{V}$ correspond to the variables while $\mathcal{E}$ is a neighborhood system (pair-wise \& higher order cliques) that encodes the relationships among these variables. In a discrete optimization problem, the aim is to assign a discrete label $l_v \in L$ to every variable $v \in \mathcal{V}$, by minimizing the following energy function:}

\REV{\begin{equation}
\label{eq:discrete}
\argmin_{l_p} \mathcal{P}(g,f) = \sum_{v\in \mathcal{V}}g_p(l_v) + \hspace{-4mm} \sum_{(v_1,v_2) \in \mathcal{E}} f_{v_1v_2}(l_{v_1},l_{v_2}) 
+ \sum_{C_i \in \mathcal{E}} f_{v_1 \ldots v_n}(l_{v_{i^1}},\ldots,l_{v_i^{\mid C_i \mid}}),
\end{equation}}

\noindent \REV{where $g_v(l_v)$ are the unary potentials, $f_{v_1v_2}(l_{v_1}, l_{v_2})$ are the pairwise terms and $f_{v_1 \ldots v_n}(l_{v_{i^1}},\ldots,l_{v_i^{\mid C_i \mid}})$ are high-order terms associated to high-order cliques $C_i$ depending on more than two variables. Note that this equation can be seen as a discrete version of Equation \ref{eq:chap3:generalSliceToVolume}, where the transformation mapping function $\varTheta$ is parameterized through the labeling $\{l_p\}$ and both similarity and regularization terms are encoded in the energy terms.}

\REV{Several models have been proposed based on the general formulation from Equation \ref{eq:discrete}}. A discrete approach to slice-to-volume rigid registration was recently proposed by \cite{Porchetto2016}. Inspired by previous works on discrete estimation of linear transformations using graphical models \citep{Zikic2010a}, they formulate it through a fully-connected pairwise MRF, where the nodes are associated to the rigid parameters, and the edges encode the relation between the variables. Deformable image registration can also be formulated as a minimal cost graph problem where the nodes of the graph correspond to the control points of a deformation grid and the graph connectivity imposes regularization constraints. Even if this technique has been applied mainly to mono-dimensional cases (2D-2D or 3D-3D) \citep{Glocker2011}, several works were published during the last years extending this theory to the case of slice-to-volume registration \citep{Ferrante2013, Ferrante2015, Ferrante2015a, Ferrante2015c}. 

The first solution to the problem of deformable slice-to-volume registration using graphical models and discrete optimization was presented by \cite{Ferrante2013}. A regular grid, which models at the same time the plane location and the in-plane deformations, is interpreted as a graphical model where every grid point corresponds to a discrete variable. A pairwise model which combines linear and deformable parameters within a coupled formulation on a 5-dimensional label space is used. The main advantage of such a model is the simplicity provided by its pairwise structure, while the main disadvantage is the dimensionality of the label space which makes inference computationally inefficient and approximate (limited sampling of search space). FastPD was adopted as the optimization algorithm given the properties of the energy function (pairwise terms and non-submodularity).  Motivated by the work of \cite{Shekhovtsov2008}, in an effort to reduce the dimensionality of the label space, the work by \cite{Ferrante2015a} presents a different model (the \REV{so-called} decoupled model) where linear and deformable parameters are now separated into two interconnected subgraphs which refer to lower dimensional label spaces. It reduces the dimensionality of the label space by increasing the number of edges and vertices, while keeping a pairwise graph. In this case, loopy belief propagation is used as optimization algorithm, since FastPD requires in general an equal number of labels for all nodes, which is an issue in their setting given the different dimensionality of the label spaces. Finally, in \citep{Ferrante2015}, a high-order approach is presented, where the label space dimensionality reduction is achieved by augmenting the order of the graphical model, using third-order cliques which exploits the expression power of this type of variable interactions. Such a model provides better satisfaction of the global deformation constraints at the expense of quite challenging inference. Loopy belief propagation is also used as the optimization algorithm in this case.

Discrete methods have several advantages when compared with continuous approaches for image registration. First, they are inherently gradient-free, while most part of continuous methods require the objective function to be differentiable. Second, continuous methods are quite often prone to be stuck in local minima when the functions are not convex. However, in case of discrete methods, even complicated functions could potentially be optimized using large neighbor search methods. Third, parallel architectures can be used to perform non-sequential tasks required by several discrete algorithms (such as message calculation in LBP) leading to more efficient implementations. Fourth, by using a discrete label space, one can explicitly control its range and resolution, what can be useful to introduce prior information. The main limitation of discrete methods when compared to continuous approaches is the accuracy, which is bounded by the discretization of the continuous space. However, as suggested by \cite{Glocker2010}, even if the optimality is bounded by the discretization, with intelligent refinement strategy the accuracy of continuous methods can be achieved.

\subsection{Miscellaneous} \label{sec:chap3:miscellaneous}
Independently of whether the variables are continuous or discrete, different strategies can be used to explore the space of solutions. Heuristics or metaheuristics can obtain acceptable solutions especially in cases where we deal with non-linear, non-convex or black-box optimization problems, even if we know that no optimality guarantee is provided. In some cases (e.g. initializing a more complex registration procedure) having an approximately good solution is enough; in these scenarios, different strategies can be envisioned.

Evolutionary algorithms are among the most popular bio-inspired metaheuristics. In evolutionary computation, the concept of biological evolution is used to explore the search space. The individuals of a given population are the candidate solutions to our problem and they evolve according to different laws (such as mutation, recombination and selection). Evaluation of the quality for a given solution is performed through a fitness function, which is in fact our objective function. The idea is that, over the generation sequence, individuals with better and better fitness are generated, leading to a good solution. These algorithms are usually stochastic. If this is the case, new candidate solutions might be sampled according to a multivariate normal distribution in the space of parameters. \REV{Genetic algorithms were used by \cite{Lin2013} to estimate a 6-DOF transformation between 2D MRI real-time slices of the knee and 3D MRI volumes.} One evolutionary algorithm applied in several slice-to-volume registration problems is the covariance matrix adaptation evolutionary strategy (CMA-ES) \citep{Ghanavati2010, Tadayyon2010, Tadayyon2011}. This method updates the covariance matrix of the aforementioned distribution, by estimating a second order model of the underlying objective function similar to the approximation of the inverse Hessian matrix. Another case based on genetic algorithms to solve our problem of interest was proposed by \citep{Gefen2005, Gefen2008}. Different optimization strategies may be combined as well depending on the parameter space. In \citep{Tadayyon2010, Tadayyon2011}, for example, the CMA-ES was applied to optimize the translation parameters of the rigid transformation model, while rotation and deformable ones were estimated using a standard gradient descent approach. In their application to prostate MR images, the CMA-ES was not able to optimize a 6-DOF search space as it diverged on rotations regardless of scaling. 

Another popular metaheuristic is the well known simulated annealing (SA) method. It mimics the physical process that metal atoms suffer when the material is heated and then slowly cooled. In order to avoid local minima, the algorithm explores new directions which lower the objective function, but also, with a certain probability, those that raise the objective. This probability decreases with the number of iterations, so that it stays away from local minima in early iterations and is able to explore globally for better solutions. \cite{Birkfellner2007} used this method to register fluroCT slice to CT volumes optimizing several standard iconic matching criteria, while \cite{Cifor2013} applied it to a multi-slice to volume registration case where a robust modality-independent similarity measure was optimized.

\section{Number of Slices} \label{sec:chap3:numberOfSlices}
In this review we include single and multi slice-to-volume registration methods. While in the first case the estimated slice-to-volume mapping function considers just a single 2D slice, in the second case several slices are mapped to the 3D volume.

A trivial extension of any single slice-to-volume registration method to the multi-slice case could be achieved by simply applying the algorithm to every input slice independently. It would allow parallelizing the process -since every registration could be performed in parallel- but, at the same time, contextual information would be lost since no relation among the input slices would be considered. Several methods of this type have been proposed. We can cite \citep{Fei2003, Ferrante2015, Osechinskiy2011a, Yavariabdi2015} among others. 

Different strategies can be adopted to register multiple slices to a given volume. Here we only review methods that perform a consistent transformation to the set of input slices, but still conceive the process as slice-based without seeing them as a complete and unique volume. Methods that reconstruct a new volume using the input slices, and then perform a standard 3D-3D registration process, are not taken into account. Note that, in some works like \citep{Brooks2008, Yan2012, Xu2014a}, experiments have been performed to compare the performance of multi slice-to-volume versus volume-to-volume registration. In \citep{Xu2014a}, the comparison was done in the context of MR prostate images, in order to show that multi slice-to-volume registration is sufficient in capturing prostate motion intra-operatively. The conclusion was that multi slice-to-volume registration was able to produce results that were close enough to volume-to-volume registration. This is an encouraging result, considering the data computation reduction achieved when using only a few slices instead of a complete volume. \citet{Brooks2008} and \citet{Yan2012} arrived to similar conclusions in different medical scenarios. This consensus indicates the reliability of using sparse slices instead of the full volume when possible. Several advantages are associated to this strategy: (i) the computational requirements are lower thanks to the reduction in the amount of data to be processed (only a few slices vs a complete volume), (ii) the omission of the potentially complex reconstruction step, (iii) greater adaptability of the similarity measures and (iv) easier parallelization \citep{Brooks2008}.

In the presence of multiple slices, multi slice-to-volume procedure -as opposed to single slice procedure- greatly increases the quality of the matching leading to more robust registration methods \citep{Yu2011}. This is even more interesting when the relative position between the slices is known. In \citep{Chandler2006, Chandler2008}, for example, multi slice-to-volume registration was used to correct for misaligned cardiac anatomy in Short Axis (SA) images by registering stacks of two parallel slices (which are supposed to be aligned between them) to a high-resolution 3D MR axial cardiac volume. In other cases, different configurations like orthogonal slices \citep{Gill2008, Leung2010, Tadayyon2011, Miao2014, Xu2014a}, slices aligned in arbitrary positions (obtained using a tracking system) \citep{Wein2008, Heldmann2009, Heldmann2010, Olesch2011, Olesch2011a, Cifor2013} or even temporal sequences of slices \citep{Miao2014}, are registered to a volume. Another case is \cite{Yu2008, Yu2011}, where a bi-protocol was proposed to reconstruct a microscopic volume of a cell. The bi-protocol is composed by two sets of multiple slices imaging the same cell, which are captured with two different geometries.
Both sets of slices are then registered to each other and a final volume is reconstructed using a simple interpolation strategy.

Let us recall that multi slice-to-volume registration involves lower computational complexity compared to 3D-3D registration methods. Furthermore, when compared to single slice methods, multi slice-to-volume registration certainly improves the robustness of the registration process by augmenting the image support. Dealing with several slices requires more computing power than the single slice case; however, on occasions like freehand US sweeps, slices contain redundant information that could be avoided. In that sense, as the complexity of the scheme is proportional to the number of input slices, \cite{Wein2008} proposed a strategy which starts out by selecting only a few key-slices. These are used to estimate a rigid or affine transformation model, mapping the US slices to a CT volume. Since neighboring frames of the freehand US sweep contain overlapping information, only a few key slices are selected. They chose the most informative slices by selecting those with the highest image entropy. \cite{Olesch2011} applied the same key-slice selection technique to the variational deformable registration framework proposed by \cite{Heldmann2009}. To this end, well distributed slices throughout the volume were considered, containing meaningful information in terms of entropy. In another work, \cite{Olesch2011a} extended this key-slice selection technique by introducing a focused registration strategy that only considers slices which are in a given region of interest (ROI). 

In \citep{Jiang2007a}, a multi slice-to-volume registration method -entitled Snapshot magnetic resonance imaging with Volume Reconstruction (SVR)- is proposed to deal with reconstruction of MRI of moving subjects. After imaging the target object of interest repeatedly (producing multiple overlapped stacks of slices of the same moving object) to guarantee sufficient sampling, one of the stacks is chosen. Then, an iterative mutli slice-to-volume registration strategy is applied where the stacks are subsequently registered (in a rigid way) to the reference. Once all the stacks are registered, they are averaged to produce a new reference. Then, the number of temporally contiguous slices in every stack is reduced so that the data is divided into subpackages that are temporally contiguous although the slices in each subpackage may not be spatially adjacent. The process is then repeated: the subpackages are registered to the average image, which is again updated, and the number of slices per stack is once more decreased; this process is repeated until each slice is treated in isolation (reducing to the single slice-to-volume case) and the final average volume is reconstructed. In \citep{Jiang2007, Jiang2009} the idea was extended to allow the reconstruction of diffusion tensor images in the same scenario. In the same work, \cite{Jiang2007a} tested the benefit of using both, parallel and orthogonally acquired slices with prospectively acquired data and simulated cases as well. 
More recently, a fast multi-GPU accelerated framework to perform SVR was presented \citep{Kainz2015}. They proposed a fully parallel SVR approach for the reconstruction of high-resolution volumetric data from motion corrupted stacks of images. Parallelization is performed at two levels: (i) at the slice level, multiple slices are treated separately for large parts of the reconstruction process and (ii) since pixel/voxel based operations are independent of each other, they can also be executed in parallel. Authors claimed that their approach is five to ten times faster than the fastest currently available multi-CPU frameworks.

\citet{Rousseau2005, Rousseau2006} proposed the registration of multiple sets of orthogonal 2D MRI slices into a high resolution MRI volume. The work of \cite{Jiang2007a} shares some similarities with them. However, instead of reducing the number of slices per subpackage until each slice is treated in isolation, a two-step approach is performed. In this setting, a global alignment of the low resolution images is followed by the registration of each slice of the low resolution images to the reconstructed high resolution volume. Only orthogonal sets of slices were considered for reconstruction. 

The work by \citet{Kim2008, Kim2010} (built on top of \citep{Rousseau2005, Rousseau2006}), proposed a new approach entitled Slice Intersection Motion Correction (SIMC). It considers the registration process directly in terms of the intersections of each pair of slices in the stacks, avoiding the intermediate volume estimation process. Independent per-slice rigid transformations are estimated by minimizing a global energy function defined by the sum of dissimilarity measures of all intersection profiles between any two orthogonal slices. SIMC can be used in this scenario, considering intersecting lines of voxels instead of standard patch or global based calculation. \citet{Kim2008, Kim2010} applied SIMC to the problem of reconstructing MR images of a moving human fetal brain. Since then, several extensions to this work have been published, like for example the work presented in \citep{Kim2010a} expanding the method to the case of Diffusion Tensor Images and \citep{Kim2010b, Kim2011} which modified it in order to account for bias field inconsistency correction on fetal brain MR images.

An approach based on particle filtering (PF) was considered in \citep{Nir2011, Nir2014} to deal with multi slice-to-volume registration of histological images to MR or US volumes. PF framework derives an optimal estimation of the parameters in a Bayesian fashion, tackling two of the main issues that arise from multimodal registration, such as susceptibility to initialization and optimal solutions. PF framework was applied to the problem of head motion tracking in the context of EPI based fMRI motion correction by \cite{Chen2016}. Chen and coworkers used this strategy to couple successive EPI slices during the registration process, resulting in improved performance when compared with standard volume-to-volume and single slice-to-volume registration approaches.

\section{Applications} \label{sec:chap3:application}
A broad number of clinical settings and applications can benefit from slice-to-volume registration. In this review we classify them in two main categories. 

\subsection{Image Fusion and Image Guided Interventions (IGI)} \label{sec:chap3:imageFusionIGI}

\begin{figure}[t!]
  \centering
     \includegraphics[width=\textwidth]{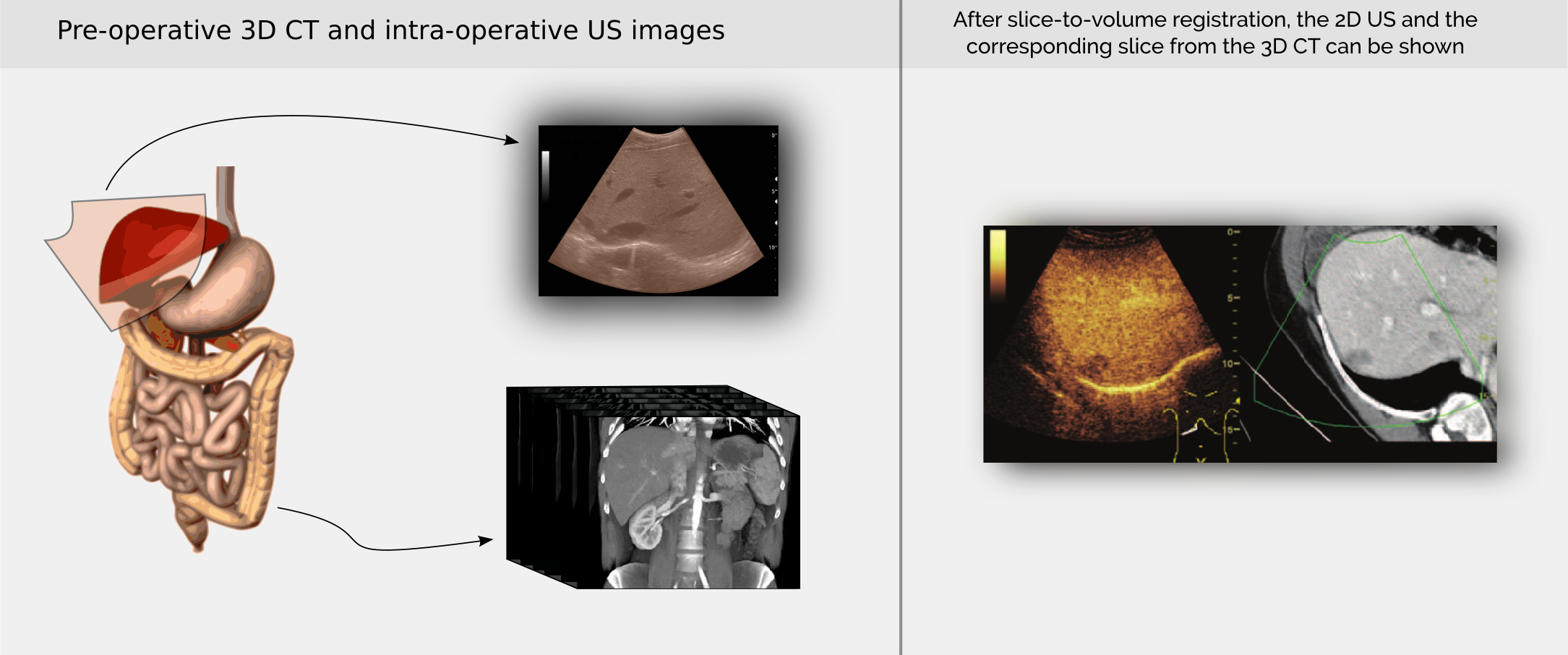}
     \caption{\REV{One of the main applications requiring slice-to-volume registration is image fusion for image guided procedures. In the figure we can observe an example where an intra-opearative US slice is registered to a pre-operative CT image. Images are then shown side-by-side, providing complementary information (The side-by-side US/CT image was extracted from \cite{Ewertsen2013}).}}
  \label{fig:imageFusion}
\end{figure}

Several medical procedures such as image guided surgeries and therapies \citep{Fei2002}, biopsies \citep{Xu2014a}, radio frequency ablation \citep{Xu2013}, tracking of particular organs \citep{Gill2008} and minimally-invasive procedures \citep{Liao2013, Huang2009} belong to this category. In this context, slice-to-volume registration brings high resolution annotated data into the operating room. Generally, pre-operative 3D images such as computed tomography (CT) or magnetic resonance images (MRI) are acquired for diagnosis and manually annotated by expert physicians. During the surgical procedure, 2D real time images are generated using different technologies (e.g. fluoroCT, US or interventional MRI slices). The alignment of intra-operative images with pre-operative volumes augments the information that physicians have access to, and allows them to navigate the volumetric annotation while performing the operation \REV{(see Figure \ref{fig:imageFusion})}. These intra-operative images inherit lower resolution and quality than the pre-operative ones. Moreover, tissue shift collapse as well as breathing and heart motion during the procedure, cause elastic deformation in the images, what makes slice-to-volume registration an extremely challenging task. A statistically significant improvement in alignment has been demonstrated when comparing automatic methods to manual (human) results, showing the importance of automatic slice-to-volume registration algorithms in the context of image fusion and IGI \citep{Fruhwald2009}.

The pioneering work of Fei and coworkers \citep{Fei2001, Fei2002, Fei2003, Fei2004a, Fei2004} introduced iconic slice-to-volume registration to the problem of image fusion in the context of image guided surgeries. The motivation was that low-resolution Single Photon Emission Computed Tomography (SPECT) can be brought to operating by pre-registering it with a high-resolution MRI volume, which could be subsequently fused with live-time iMRI. That is how, by registering the high-resolution MR image with live-time iMRI acquisitions, Fei and coworkers could map the functional data and high-resolution anatomic information to live-time iMRI images for improved tumor targeting during thermal ablation. 

\cite{Birkfellner2007} used slice-to-volume registration to fuse 2D fluoroCT with volumetric CT, which is a well know tool for image-guided biopsies in interventional radiology. In this case, the pre-interventional diagnostic high resolution CT with contrast agent is used to localize a lesion in the liver. However, during the intervention, the lesion is no longer visible. Thus, localizing the slice of the CT that corresponds to the intra-operative fluoroCT allows doctors to find the lesions during the biopsy. This approach only considers rigid transformations. However, interventional procedures like radio frequency ablation (RFA) or image-guided biopsies, which use fluoroCT as image guiding technology, are performed while the patient is breathing continuously. Therefore, deformations should also be taken into account when registering with the pre-operative static CT image. The influence of such deformations and the reliability of performing non-rigid registration in such scenario was discussed in \citep{Micu2006, Lasowski2008}. It was claimed that a 2D-3D non-rigid registration solution -based on the single low quality fluoroCT- cannot be precise as required to performed medical procedures. This is mainly due to the poor support in terms of liver anatomical features (mainly vessels) provided by the fluoroCT slices. They proposed to overcome this limitation by providing an adaptive visualization \citep{Lasowski2008} of the volume area surrounding the minimum estimated pose. This approach copes with the uncertainty in estimating the deformation and brings more information than a single registered slice. Their method performs rigid slice-to-volume registration, and includes views of the CT-Volume determined along flat directions of the out-of-plane motion parameters next to the minimum pose. 

Laparoscopic and endoscopic interventional procedures also exploit slice-to-volume registration. \cite{SanJoseEstepar2009}, proposed a method to register endoscopic and laparoscopic US images with pre-operative computed tomography volumes in real time. It is based on a new phase correlation technique called LEPART accounting for rigid registration. Other methods applying slice-to-volume registration on minimally invasive procedures can be found here \citep{Heldmann2010, Bao2005}.

\subsection{Motion Correction and Volume Reconstruction} \label{sec:chap3:motionCorrectionVolumeReconstraction}
  
\begin{figure}[t!]
  \centering
     \includegraphics[width=\textwidth]{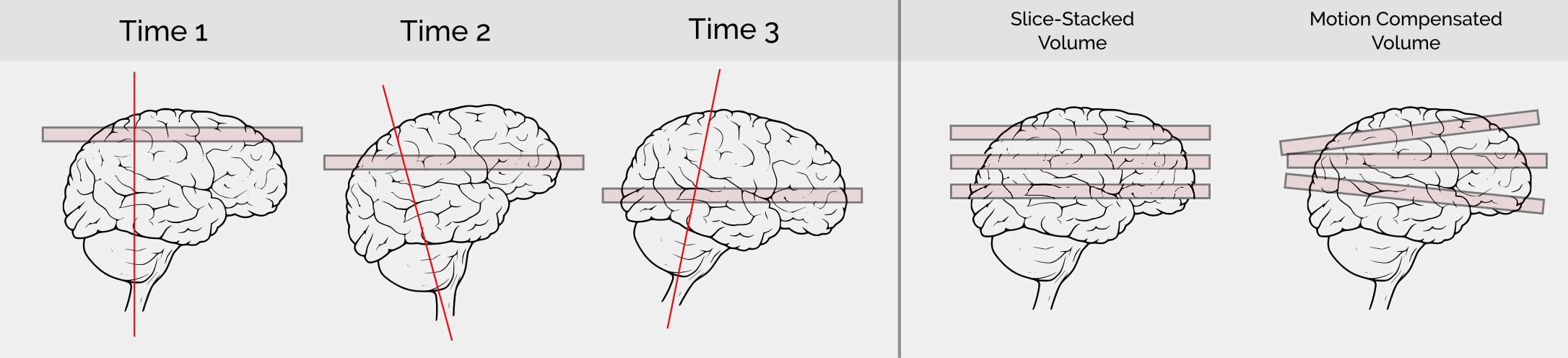}
     \caption{\REV{Slice-to-volume registration is a key task when performing motion correction for volume reconstruction. In the figure we can observe a typical case, where the patient moves while the 3D image is being acquired slice-by-slice, resulting in a corrupted slice-stacked volume. Motion correction through slice-to-volume registration is a common technique to improve the quality of the reconstructed volume (this figure is based on Figure 1 from \cite{Chen2016}).}}
  \label{fig:motionCorrection}
\end{figure}

The second category is motion correction and volume reconstruction. Here, the goal is to correct misaligned slices when reconstructing a volume of a certain modality \REV{(see Figure \ref{fig:motionCorrection})}. A typical approach to solve this task consists in mapping individual slices within a volume onto another reference volume in order to correct the inter-slice misalignment. The popular map-slice-to-volume (MSV) method introduced this idea \citep{Kim1999}. More recently, applications of slice-to-volume registration to the same problem in different contexts like cardiac magnetic resonance (CMR) \citep{Chandler2008, Elen2010}, fetal images \citep{Seshamani2013} and diffusion tensor imaging (DTI) \citep{Jiang2009} have shown promising results. In these problems, it is usually assumed that a single slice is coherent, in the sense that spatial inconsistency only happens at the inter-slice level.

Slice motion correction in the context of volume reconstruction, typically involves iterative registration of slices to a target volume. The target volume may be an anatomical reference or it could be reconstructed at each iteration using current estimates of slice motion, considering all possible views of the subject.

Kim and coworkers \citep{Kim1999} introduced the map-slice-to-volume (MSV) approach for inter-slice motion correction in time series of fMRI image. In such cases, head motion represents the major source of error in measuring intensity changes related to given stimuli in fMRI time series \citep{Yeo2004}. The aim of MSV is to retrospectively remap slices that are shifted by head motion to their spatially correct location using an anatomical MRI volume as reference. The MSV approach estimates 6-DOF rigid transformations independently for every fMRI slice, by minimizing the energy based on mutual information, using a Nelder-Mead downhill simplex method for optimization. This method was presented as an alternative to the slice-stack approach. Instead of considering slice-wise registration, it assumes stacks of slices being already registered among them, and ignores the inter-slice motion inherent to multi slice echo planar imaging (EPI) acquisition sequence (since each slice is excited at a sequential time interval). The MSV showed better performance than previous volume-to-volume registration methods, but the reliability of the estimated position parameters for the end cap slices was low due to the limited information support of smaller regions (less textured area and more background). \cite{Park2004} proposed to overcome this limitation using Joint Mapping of Slices into Volume (JMSV). JMSV is a multi-slice registration method that jointly estimates a rigid body transformations per slice, while penalizing the implied acceleration in the motion trajectory of the subject -i.e. abrupt changes in the motion parameters of sequentially acquired slices. Other extensions to the standard MSV include (i) accounting for deformable registration \citep{Kim2000} through TPS transformation models; (ii) improving motion correction capability of MSV with concurrent iterative field-corrected reconstruction \citep{Yeo2004, Yeo2008}; (iii) extensive evaluation of the activation detection performance of MSV and effects of temporal filtering of motion parameter estimates \citep{Yeo2006} and spin saturation effect \citep{Kim2008a}; (iv) improving the MI matching criterion estimation in the low-complexity end-slices (slices near top or bottom of the head scans, presenting poorer information) by incorporating joint probability density functions of image intensities estimated from successfully registered center-slices in the same time-series.   

After the fundamental work of \cite{Kim1999} and its extensions (most of them related to fMRI image reconstruction), another problem requiring to correct for inter-slice motion started to attract attention from the community: fetal brain MR imaging. In this case, there is no anatomical reference available to be used as a target volume; therefore, a reference volume is calculated during the registration process using the current estimate of the slices. \cite{Jiang2007} and \cite{Rousseau2006} established the basis of a new family of methods which rely on iterations of slice-to-volume registration and scattered data interpolation (SDI) to perform super-resolution reconstruction of moving objects -in particular, fetal brain MRI-. Both approaches share a similar and iterative slice-to-volume registration scheme, but differ on the SDI method: while \cite{Rousseau2006} use Gaussian kernel-based SDI, \cite{Jiang2007} relied on a regular grid of control-points that acts as a cubic B-spline to perform SDI. In a more recent work, \cite{Gholipour2009, Gholipour2010} criticized SDI-based approaches stating that none of these techniques can guarantee the convergence of the reconstruction to, at least, a local optimal solution. Thus, they proposed the use of a maximum likelihood (ML) error minimization approach, which guarantees the convergence of the reconstructed image to match the motion-corrected slices through steepest descent error minimization. An extension to this method was presented in \citep{Gholipour2011}, using a real-time sensor based tracking (i.e. a non-image based approach) to estimate an initial position of every slice, which was then refined through a retrospective slice-to-volume registration approach based on ML error minimization.

\section{Validation of Slice-to-Volume Registration Methods}
\label{sec:validation}
\REV{A main issue identified during this study is the lack of open/public benchmarks with gold standard annotations specially conceived to validate slice-to-volume registration methods. This type of datasets, generally available for most medical image analysis problems\footnote{\REV{An extensive list of datasets for benchmarking several biomedical image analysis tasks can be found in the Grand Challenge website (https://grand-challenge.org/). Note that none of these datasets are specially conceived for slice-to-volume registration.}}, enable fare comparison of new approaches with the state-of-the-art, leading to more constructive contributions. One of the few public datasets used in several slice-to-volume registration studies \citep{Rivaz2014a,Rivaz2014b,Rivaz2014c,Ferrante2013,Pardasani2016} is the MNI BITE (Montreal Neurological Institute's Brain Images of Tumors for Evaluation database) introduced by \citet{Mercier2012}. It contains 2D US, 3D US and MRI brain images of several patients, with ground truth provided in form of anatomical landmarks. Even if this dataset was used in several studies, it was not specially designed to validate slice-to-volume registration methods, particularly not in case of single slice methods. As stated by \citet{Pardasani2016}, one caveat is that these landmarks were identified in the reconstructed US volumes (not in the independent slices), and therefore different workarounds (see Section \ref{sec:validation:landmarks}) have to be used to adapt it to single slice-to-volume registration validation (where we need homologous annotations in both, 2D and 3D images).}

\REV{New validation datasets specifically designed to evaluate slice-to-volume registration methods should include annotations in form of anatomical landmarks, manual segmentation of structures of interest and gold standard transformations mapping slice-to-volume in both, 2D and 3D images. Public datasets for benchmarking proved to be essential to build communities around vision problems (like Pascal VOC dataset \citep{Everingham2010} for object class recognition or IBSR dataset \citep{ibsr} for brain image segmentation) and they would certainly contribute to make slice-to-volume registration a well established problem.}

\REV{In what follows, we analyze the standard approaches that have been used to validate the methods considered in this review, discussing advantages and disadvantages in the context of slice-to-volume registration.} 

\subsection{\REV{Type of images}}
\REV{There are two main type of images used for validation: phantoms and clinical images. Phantoms are artificial objects which can be scanned under controlled conditions to simulate the acquisition of real images of a patient. They are usually designed to mimic certain properties of human or animal tissue, and used to calibrate the parameters of medical imaging devices or for benchmarking purposes. Not only artificial anatomical structures (like breast \citep{Marami2011}, heart \citep{Huang2009}, abdominal structures \citep{Yang2015, Birkfellner2007}, lumbar spine \citep{Yan2012}, pelvic organs \citep{Fei2004}, etc) but also other type of non conventional objects (like Lego bricks \citep{Xiao2016} or even pineapples \citep{Gholipour2011}) have been considered as phantoms to be imaged in the context of slice-to-volume registration. Moreover, digitally simulated phantoms (like the BrainWeb \citep{Cocosco97brainweb} considered in \citet{Gholipour2010} or the well-known analytical Shepp-Logan phantom \citep{shepp1974fourier} adopted by \citet{Kim2011}) have also been used. One of the main advantages of digitally simulated phantoms is that, since we known in detail how the image was generated, we can provide extra information like segmentation masks or probability maps for certain structures of interest, which can then be used for validation as we will see in Section \ref{sec:validationStrategies}. The other type of images that can be considered are clinical studies. This is the most common and valuable case, where images of real patients are captured and used to validate a registration method.}

\REV{Most of the methods considered in this review use clinical images for validation, since it proves the potential of the method to be applied in real clinical context. However, it must be noted that one of the main drawbacks of clinical datasets is the lack of annotations. Finding annotations for clinical images is harder than in case of phantoms, since we usually require specialists who can manually annotate them. Therefore, a common strategy followed by several authors like \cite{Xu2008, Huang2009, SanJoseEstepar2009, Kim2011, Gholipour2011,Schulz2014, Kainz2015} is to combine phantom and clinical datasets, in a effort to get the best of both worlds: accurate measurements based on phantom data, and realistic validation using clinical data.}

\subsection{\REV{Measuring the Performance of Slice-to-Volume Registration Methods}}
\label{sec:validationStrategies}
\REV{Alternative indicators have been considered to quantify the performance of slice-to-volume registration. While some of them measure global properties (like the estimated parameters in rigid registration) other methods focus on local aspects (for example, the error in specific anatomical landmarks or local differences in terms of image intensity before and after registration). Table \ref{table:validationMetrics} includes a summary of the most relevant metrics used to validate the methods included in this survey. Here we discussed them in detail.}

\subsubsection{\REV{Distance Between Transformation Parameters}}
\REV{Given a single (or multiple) slices and a volume, if the transformation $\varTheta_{GT}$ that maps both images is known, we can estimate the distance between $\varTheta_{GT}$ and the estimated transformation $\hat{\varTheta}$. This approach is mostly used to validate global linear transformations (see for example \cite{Kim1999, Park2004, Dalvi2008, Ferrante2013, Lin2013}) where the number of parameters to estimate is low and distances per parameter $\mid \varTheta^i_{GT} - \hat{\varTheta^i} \mid$ can be reported. However, in \cite{Su2013}, the same strategy is applied to measure the accuracy of the estimated deformation field. In that context, they simply consider the average of the euclidean distances over the components of the deformation field in every pixel, given by $\frac{1}{N} \sum_{x \in \Omega} \mid \mid d_{GT}(x) - \hat{d}(x) \mid \mid$ where $d_{GT}(x), \hat{d}(x)$ are the corresponding displacement vectors of the ground-truth and estimated deformation fields in the position $x$ respectively, $N$ is the total number of pixels and $\Omega$ is the image domain. }

\REV{Note that this strategy requires to know a priori the exact transformation (rigid or non-rigid) $\varTheta_{GT}$ mapping slice-to-volume. In both phantom and clinical scenarios, this is rarely the case. To overcome this limitation, a common strategy applied in several works (e.g. \citet{Kim1999, Park2004, Yeo2004, Smolikova2005, Yeo2006,Birkfellner2007, Dalvi2008, Gefen2008, Kim2008a, Yeo2008, Elen2010, Gholipour2010, Zakkaroff2012, Ferrante2013, Ferrante2015a, Ferrante2015, Porchetto2016}) consists in generating synthetic ground truth by extracting arbitrary slices from known transformations given a 3D image. These parameters are then perturbed to initialize the registration process, and used as ground-truth to calculate the error with respect to the estimated transformation. Another alternative is to generate bronze-standard annotations. In this case, the idea is to use more reliable registration methods when possible (e.g. volume-to-volume registration when multiple slices are available, or landmark based registration when accurate anatomical landmarks are provided) to obtain the transformations, which are then considered as ground-truth. Alternative methods to produce bronze standard annotations for slice-to-volume registration can be found in \cite{Fei2002, Fei2003, Fei2004, Fei2004a,Penney2006,Jiang2009}.}

\subsubsection{\REV{Point-based Registration Error}}
\label{sec:validation:landmarks}
\REV{Another common strategy frequently used in the literature is based on landmarks. The idea is to annotate points of interest which are visible in both, the slice and the volume images, so that we can measure the distance between the corresponding points before and after registration. The mean distance between the ground truth and registered anatomical landmarks is commonly referred as target registration error (TRE), and calculated as:
\begin{equation}
\sqrt[]{\frac{1}{M} \sum_{k=1}^{M} \mid \mid p^k_{GT} - T_{\hat{\varTheta}} \circ p^k \mid \mid},
\end{equation}
\noindent where $p^k, p^k_{GT}$ are the corresponding landmarks in the source and target domains, $M$ is the number of landmarks and $T_{\hat{\varTheta}}$ is the estimated transformation after registration parameterized by $\hat{\varTheta}$. This coefficient is sometimes referred as root-mean-squared error (RMS).
If fiducial markers are used instead of anatomical points, then the metric is referred as fiducial registration error (FRE). Both concepts were first introduced in the seminal work of \cite{fitzpatrick1998predicting} to measure the performance of a geometric registration method, but have been adopted by the registration community as a general validation metric. Different variations of TRE have been used to validate slice-to-volume registration methods in \cite{Bao2005,Rousseau2005,Penney2006,Rousseau2006,Hummel2008,Leung2010,Wein2008,Huang2009,Tadayyon2011,Yan2012,Cifor2013,Fuerst2014,Nir2014,Rivaz2014a,Rivaz2014b,Rivaz2014c,Xu2014a,Xu2014b,Yavariabdi2015,Xiao2016}, including fiducial and anatomical landmarks.}

\REV{Landmark-based error is a reliable indicator of the registration quality. However, it must be noted that in slice-to-volume registration cases where slices are too sparse, annotating such points of interest can be extremely difficult and sometimes impossible. A simple workaround was adopted by \cite{Pardasani2016} to adapt the MNI BITE dataset (where landmarks are identified in the reconstructed US volumes and the 3D MRI images) to the slice-to-volume registration case: they only used the US slices that were within 0.3mm of an expert-identified landmark. This reduced the number of available slices, but allowed them to compute the TRE indicator.}

\REV{An alternative measurement based on point distances can be adopted even if landmarks are not available. In this case, the position of the voxels within a region of interest is compared with the corresponding voxels in the target before and after registration. Approaches considering distances between voxels have been used in several slice-to-volume registration methods like \cite{Fei2002,Fei2004,Fei2004a,Fei2003,Chandler2006,Kim2008,Tadayyon2010,Kuklisova-Murgasova2012,Fogtmann2014}. However, we still require a way to establish correspondences between the voxels in both images, which is not always available.} 

\subsubsection{\REV{Segmentation-based Metrics}}
\REV{When segmentation masks are available for certain structures of interest in both slice and volume images, segmentation-based statistics computed before and after registration can be used for validation. The rational is that structures that were not aligned before registration, should be afterwards. Thus, the usual strategy consists in extracting a slice from the volume at the initial position, and doing the same after registration. At this point, the problem reduces to computing segmentation-based statistics between two 2D images. Alternative segmentation-based coefficients have been considered to deal with slice-to-volume registration validation. }

\REV{Dice similarity coefficient \citep{dice1945} (DSC, also known as S{\o}rensen-Dice coefficient) quantifies the amount of overlapping between two given segmentations masks A and B following $DSC(A,B) = \frac{2 \mid A \cap B \mid}{\mid A \mid + \mid B \mid}$. Its value ranges from 0, meaning no spatial overlap, to 1, indicating complete overlap. This coefficient has been widely used in several slice-to-volume registration methods, where segmentation masks of liver tumor \citep{Cifor2013}, brain structures \citep{Ferrante2013,Ferrante2015,Ferrante2015a,Kainz2015}, prostate \citep{Nir2014}, cardiac left ventricle \citep{Xu2014b} and pelvic organs \citep{Yavariabdi2015} were considered. Alternative overlapping measures were used in slice-to-volume studies by \citep{Brooks2008,Chandler2008,Osechinskiy2011,Museyko2015}. As stated by \citep{Yavariabdi2015}, note that a high Dice does not mean a good contour overlap, which is a desired property after registration of structures of interest. In this case, indicators like Hausdorff distance (maximum distance between contours) and contour mean distance (CMD) can reported (examples in the case of slice-to-volume registration can be found in \cite{Sun2007,Brooks2008,Elen2010,Tadayyon2010,Tadayyon2011,Zakkaroff2012,Nir2014,Yavariabdi2015}).}

\REV{The main advantage of using segmentation masks for slice-to-volume registration validation is their availability: annotations in the form of segmentation masks are more frequent than landmarks or ground-truth transformations. Moreover, if we aim to evaluate non-rigid registration, the estimated deformation fields can be used to warp the segmentation masks, enabling quantification for deformable registration error. However, as discussed in \cite{Rohlfing2012}, attention must be given to the structures of interest used for validation. Statistics computed using wide and non well-localized structures segmentation (like brain tissue for example), may not provide sufficient evidence to validate registration accuracy. According to the authors, only smaller, more localized regions of interest which approximate point landmarks, can provide such an evidence, since their overlap approximates point-based registration error. Combining area overlapping and contour distance based coefficients (which provide complementary information about the registration quality) has been suggested as a way to alleviate such inconsistencies \cite{ferrante2015graph,Yavariabdi2015}.}

\subsubsection{\REV{Appearance-based metrics}}
\REV{Appearance information can also be used for validation. When the slice-to-volume registration method is an intermediate step towards performing motion correction/volume reconstruction, an indirect way to evaluate the registration quality is by quantifying the reconstruction accuracy. In this case, reconstruction error measured through intensity differences (usually the root mean square differences or RMSD) or signal-to-noise ratio (SNR) indicators between estimated and ground-truth volumes were used \citep{Zarow2004,Noble2005,Chandler2006,Jiang2007,Gholipour2009,Gholipour2010,Honal2010,Gholipour2011, Yu2011,Kuklisova-Murgasova2012,Kainz2015}. An alternative considered in several works \citep{Gholipour2009,Gholipour2010,Gholipour2011} is to measure the sharpness of the reconstruction. The intuition is that when uncorrected motion or error residuals between the input image acquisitions are present, the average image will be an out-of-focus motion blurred version of the imaged structure \citep{Gholipour2009}. One of the main advantages of such measure is that it does not require ground-truth reconstructed images, since it is computed directly on the estimated volume.}

\REV{Another case where simple intensity differences or correlation metrics between the estimated and target slice can be used is when performing monomodal slice-to-volume registration. In this case SAD, SSD or CC can be used to measure registration accuracy through visual error quantification since slice and volume intensities tend to be linearly correlated (see for example \cite{Marami2011,Porchetto2016}). In multimodal cases, more complex metrics (e.g. MI or NMI) or even ad-hoc similarity measures defined for specific modalities (like the $LC^2$ metric used by \cite{Wein2008a}) can also be adopted for validation.}

\REV{The last strategy consists in quantified visual inspection, where one or multiple experts are asked to visually rate the quality of the registration assigning a score according to a given scale. This approach was taken in \cite{Birkfellner2007, Boer2007,Fruhwald2009,Kim2010,Su2013}. The main disadvantage of such strategy is that it is highly time consuming and require human intervention. However, this approach can be combined with any other listed in this survey, since it provides complementary information obtained from expert knowledge, which is difficult to quantify trough objective/mathematical metrics.}



\section{Discussion and future directions} \label{sec:conclusions}
In this survey we reviewed and discussed some of the most important works in the literature related to slice-to-volume registration. These papers were classified according to different principles including matching criterion, transformation model, optimization method, number of slices, application and \REV{validation strategy}. 

One of the conclusions that emerges when analyzing Table \ref{table:comparison} (which presents a compact summary of this work) is that the majority of the methods discussed in this survey focus on rigid registration, using iconic matching criteria and continuous optimization techniques. It is, therefore, worthwhile to wonder why this tendency is so clear and marked when dealing with slice-to-volume registration. 

The fact that most of the works estimate rigid transformations may be related to two main reasons. The first one has to do with simple requirements from the application viewpoint. As stated in section~\ref{sec:chap3:rigid}, such a basic transformation model is expressive enough to explain simple slice-to-volume mapping, since it can deal with in-plane and out-of-plane translations and rotations. In clinical scenarios that do not inherit image distortion -like basic cases of inter-slice motion correction- this type of models may be enough. In more complex cases like image guided surgeries, it is a \REV{matter of fact} that intra-operative images are deformed with respect to the pre-operative volume, due to tissue shift collapse and breathing/heart motion during the procedure. \REV{Moreover, not only in intra-operative cases but also during non-interventional imaging, many body organs motion (e.g. cardiac, lung or tongue) have natural elastic motion.} Even so, rigid models still dominate in the literature. This is related to the second reason we identified: the widespread distrust of the non-rigid transformations among the physicians. From a medical perspective, it is sometimes preferable to supply reliable but approximate cues than unrealistic solutions. Non-rigid and elastic models might lead to solutions which are correct from a geometrical point of view but they are not anatomically meaningful. Further research on realistic deformation models reflecting physical tissue properties will certainly improve the accuracy of our estimations and develop trust in our methods inside the medical community.

Iconic matching criteria (based on alternative similarity measures) turned out to be the choice of preference to describe the quality of the solutions. This is somehow unexpected if we consider the lack of image support naturally associated to slice-to-volume registration, and the image noise frequently present in intra-operative, real-time and low-quality modalities, normally corresponding to the input 2D image, which makes it difficult to explain image similarities solely based on intensity information. In that sense, we identify two strategies helping to alleviate this limitation. The first one is to use (when possible) multiple slices instead of a single one. In that way, one can improve the matching quality by augmenting the image support, while keeping lower computational complexity when compared to standard 3D-3D registration methods (as explained in Section \ref{sec:chap3:numberOfSlices}, this strategy proved to be particularly useful when dealing with motion correction for image reconstruction). The second one is to combine iconic matching criteria with geometric or sensor-based strategies (as discussed in section \ref{sec:chap3:hybrid}). While the last ones are more robust to image changes and discontinuities, the former ones contribute to produce more accurate solutions. When used in tandem with sensor based methods (for example in image guided interventions where sensors can be attached to imaging devices or surgery tools), iconic similarity measures showed to be useful to refine initial rigid estimations and provide more precise results.

In terms of optimization strategy, even if the communities of computer vision and medical imaging started to massively adopt discrete approaches during the last decade, the same did not happen on the particular case of slice-to-volume registration. While most of the published methods adopt continuous optimizers (gradient and non-gradient based), only a few works recently published started to conceive this problem from a discrete perspective. Continuous methods are well established and showed to be good enough to deal with basic slice-to-volume registration problems. Nevertheless, results obtained in other biomedical image analysis problems modeled through discrete methods \citep{Paragios2016} suggest that these strategies have a great potential which remains to be exploited for slice-to-volume registration. In cases where the number of parameters to be estimated is too high (e.g deformable registration where dense deformation fields are adopted as transformation models), the energy functions are not convex (e.g. multimodal image registration) or the search space is remarkably wide (e.g. poor initializations), discrete methods may make a difference.

The research community of computer vision and medical imaging has made major efforts towards developing more accurate slice-to-volume registration strategies, mostly considering the geometric aspects of this problem. However, the massive production of image data that we are witnessing during the 21st century, combined with the latest advances in artificial intelligence and, in particular, deep convolutional networks \citep{lecun2015}, open the possibility to conceive slice-to-volume image registration within an entirely different paradigm. Deep convolutional networks trained using massive amounts of data outperformed all existent state-of-the-art strategies in other fundamental vision tasks like image segmentation \citep{long2014fully} and object detection \citep{Szegedy2013}. Some recent works on image registration and vector flow estimation using CNNs (where not only the similarity measure \citep{Simonovsky2016, Zagoruyko2015} but also the actual registration process \citep{Fischer2015, Miao2016a, Yang2016} is learned from examples) suggest that these problems are not an exception. \REV{The formulation of slice-to-volume image registration under this paradigm has just started to be explored and remains largely unexplored. In that sense, the promising results already achieved \citep{Hou2017} suggest that shifting to a learning-based slice-to-volume registration paradigm may lead to faster and more accurate predictions, opening new doors in the field.}

\section*{Acknowledgments}
This research was partially supported from the European Research Council Grant 259112.

\begin{landscape}
\begin{table}[]
\centering
\caption{\REV{Alternative validation metrics considered in the context of slice-to-volume registration.}}
\label{table:validationMetrics}
\begin{tabular}{@{}p{2.5cm}p{2cm}p{15cm}@{}}
\toprule
\rowcolor[HTML]{EFEFEF} 
\multicolumn{2}{c}{\cellcolor[HTML]{EFEFEF}\textbf{Metric}} & \hspace{7cm}\textbf{Paper} \\ \midrule
\rowcolor[HTML]{FFFFFF} 
\multicolumn{2}{p{5cm}}{\cellcolor[HTML]{FFFFFF}\REV{Distance between transformation parameters}} & \REV{\cite{Kim1999,Park2004,Yeo2004,Smolikova2005,Dalvi2008,Gefen2008,Kim2008a,Yeo2008,Fruhwald2009,Jiang2009,SanJoseEstepar2009,Elen2010,Gholipour2010,Zakkaroff2012,Ferrante2013,Lin2013,Seshamani2013,Su2013,Schulz2014,Ferrante2015,Ferrante2015a,Porchetto2016, Chen2016}} \\ \midrule
\rowcolor[HTML]{FFFFFF} 
\multicolumn{2}{c}{\cellcolor[HTML]{FFFFFF}\REV{Point-based metrics}} &  \REV{\cite{Fei2002,Fei2003,Fei2004,Fei2004a,Bao2005,Rousseau2005,Chandler2006,Penney2006,Rousseau2006,Kuklisova-Murgasova2012,Kim2008,Hummel2008,Leung2010,Wein2008,Huang2009,Tadayyon2010,Tadayyon2011,Yan2012,Cifor2013,Chicherova2014,Fogtmann2014,Fuerst2014,Nir2014,Rivaz2014a,Rivaz2014b,Rivaz2014c,Xu2014a,Xu2014b,Hallack2015,Yavariabdi2015,Xiao2016,Chen2016}}\\ \midrule
\rowcolor[HTML]{FFFFFF} 
\cellcolor[HTML]{FFFFFF} & \REV{Overlapping error} & \REV{\cite{Brooks2008,Chandler2008,Osechinskiy2011,Cifor2013,Ferrante2013,Museyko2015,Nir2014,Xu2014b,Ferrante2015a,Kainz2015,Yavariabdi2015}} \\ \cmidrule(l){2-3} 
\rowcolor[HTML]{FFFFFF} 
\multirow{-2}{*}{\cellcolor[HTML]{FFFFFF}\REV{Segmentation}} & \REV{Contour distances} & \REV{\cite{Sun2007,Brooks2008,Elen2010,Tadayyon2010,Tadayyon2011,Zakkaroff2012,Nir2014,Ferrante2015a,Yavariabdi2015}} \\ \midrule
\rowcolor[HTML]{FFFFFF} 
\cellcolor[HTML]{FFFFFF} & \REV{Intensity / reconstruction} & \REV{\cite{Zarow2004,Noble2005,Chandler2006,Jiang2007,Wein2008a,Gholipour2009,Osechinskiy2009,Gholipour2010,Honal2010,Gholipour2011,Marami2011,Yu2011,Kuklisova-Murgasova2012,Su2013,Kainz2015,Porchetto2016,Marami2016a,Marami2016b}} \\ \cmidrule(l){2-3} 
\rowcolor[HTML]{FFFFFF} 
\multirow{-3}{*}{\cellcolor[HTML]{FFFFFF}\REV{Appearance}} & \REV{Visual inspection} & \REV{\cite{Birkfellner2007,Boer2007,Fruhwald2009,Kim2010,Su2013}} \\ \bottomrule
\end{tabular}
\end{table}
\end{landscape}
\section*{References}

\bibliographystyle{model2-names}
\bibliography{library}






\end{document}